\documentclass[conference]{IEEEtran}
\IEEEoverridecommandlockouts

\usepackage{amsmath,amssymb,amsfonts}
\usepackage{graphicx}
\usepackage{textcomp}
\usepackage{xcolor}
\usepackage{booktabs}
\usepackage{multirow}
\usepackage{enumitem}
\usepackage{caption}
\usepackage{float}
\usepackage{algorithm}
\usepackage{balance}
\usepackage{amsthm}
\usepackage{amsfonts}
\usepackage{subfigure}
\usepackage{tikz}
\usepackage{pgfplots}
\usepackage[noend]{algpseudocode}
\usepackage{hyperref}
\hypersetup{
colorlinks=true,
linkcolor=black
}
\usepackage[numbers,sort&compress]{natbib}
\newtheoremstyle{definitionstyle}
  {0.5\baselineskip} 
  {0.5\baselineskip} 
  {\normalfont} 
  {} 
  {\bfseries} 
  {.} 
  {.5em} 
  {} 
\theoremstyle{definitionstyle}
\newtheorem{mydef*}{Definition}

\def\BibTeX{{\rm B\kern-.05em{\sc i\kern-.025em b}\kern-.08em
    T\kern-.1667em\lower.7ex\hbox{E}\kern-.125emX}}

\DeclareRobustCommand*{\IEEEauthorrefmark}[1]{%
  \raisebox{0pt}[0pt][0pt]{\textsuperscript{\footnotesize\ensuremath{#1}}}}

\title{Learning Time-aware Graph Structures for Spatially Correlated Time Series Forecasting}

\author{
  \IEEEauthorblockN{Minbo Ma\IEEEauthorrefmark{1}, Jilin Hu\IEEEauthorrefmark{2}, Christian S. Jensen\IEEEauthorrefmark{3}, Fei Teng\IEEEauthorrefmark{1}\IEEEauthorrefmark{*}\thanks{Corresponding authors: Fei Teng and Peng Han.}, Peng Han\IEEEauthorrefmark{4}\IEEEauthorrefmark{*}, Zhiqiang Xu\IEEEauthorrefmark{5}, Tianrui Li\IEEEauthorrefmark{1}}
  
  \IEEEauthorblockA{
  \IEEEauthorrefmark{1}School of Computing and Artificial Intelligence, Southwest Jiaotong University, China,\\
  \IEEEauthorrefmark{2} East China Normal University, China,
  \IEEEauthorrefmark{3}Aalborg University, Denmark,  \\ 
  \IEEEauthorrefmark{4}University of Electronic Science and Technology of China, China,\\
  \IEEEauthorrefmark{5}Mohamed bin Zayed University of Artificial Intelligence, United Arab Emirates\\
    \{minboma@my.swjtu.edu.cn, jlhu@dase.ecnu.edu.cn, csj@cs.aau.dk, \\
    \{fteng,trli\}\@@swjtu.edu.cn, penghan\_study\@@foxmail.com, zhiqiang.xu@mbzuai.ac.ae\}}
}

\begin{document}
\maketitle
\thispagestyle{plain}
\pagestyle{plain}

\begin{abstract}
Spatio-temporal forecasting of future values of spatially correlated time series is important across many cyber-physical systems (CPS). Recent studies offer evidence that the use of graph neural networks to capture latent correlations between time series holds a potential for enhanced forecasting. However, most existing methods rely on pre-defined or self-learning graphs, which are either static or unintentionally dynamic, and thus cannot model the time-varying correlations that exhibit trends and periodicities caused by the regularity of the underlying processes in CPS. To tackle such limitation, we propose Time-aware Graph Structure Learning (TagSL), which extracts time-aware correlations among time series by measuring the interaction of node and time representations in high-dimensional spaces. Notably, we introduce time discrepancy learning that utilizes contrastive learning with distance-based regularization terms to constrain learned spatial correlations to a trend sequence. Additionally, we propose a periodic discriminant function to enable the capture of periodic changes from the state of nodes. Next, we present a Graph Convolution-based Gated Recurrent Unit (GCGRU) that jointly captures spatial and temporal dependencies while learning time-aware and node-specific patterns. Finally, we introduce a unified framework named Time-aware Graph Convolutional Recurrent Network (TGCRN), combining TagSL, and GCGRU in an encoder-decoder architecture for multi-step spatio-temporal forecasting. We report on experiments with TGCRN and popular existing approaches on five real-world datasets, thus providing evidence that TGCRN is capable of advancing the state-of-the-art. We also cover a detailed ablation study and visualization analysis, offering detailed insight into the effectiveness of time-aware structure learning.
\end{abstract}


\section{Introduction}
Cyber-physical systems (CPS) that are capable of responding dynamically to real-time changes in the physical world based on input from sensors hold many benefits. In this setting, the forecasting of these time series produced by spatially distributed sensors plays an essential role, as it allows CPS to make informed decisions and dynamically adjust to the ever-changing physical world. This ultimately enables improved overall efficiency, reliability, and responsiveness in a wide range of domains, such as air quality forecasting~\cite{zheng2015forecasting}, weather forecasting~\cite{lin2022conditional}, transportation planning~\cite{morsali2020spatio}, and vessel collision risk warning~\cite{zhen2017novel}.

\begin{figure}[]
	\centering
	\subfigure[Spatial correlations between stations located in different functional areas]{
		\begin{minipage}[b]{0.8\columnwidth}
                \centering
			\includegraphics[width=\columnwidth]{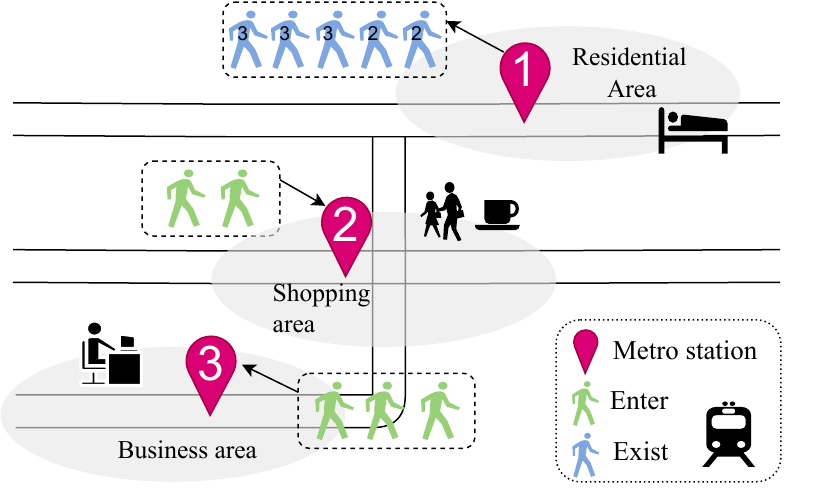}
		\end{minipage}
		\label{fig:motivation1}
	}
        \subfigure[Spatial trend and spatial periodicity]{
            \begin{minipage}[b]{\columnwidth}
                \centering
            \includegraphics[width=\columnwidth]{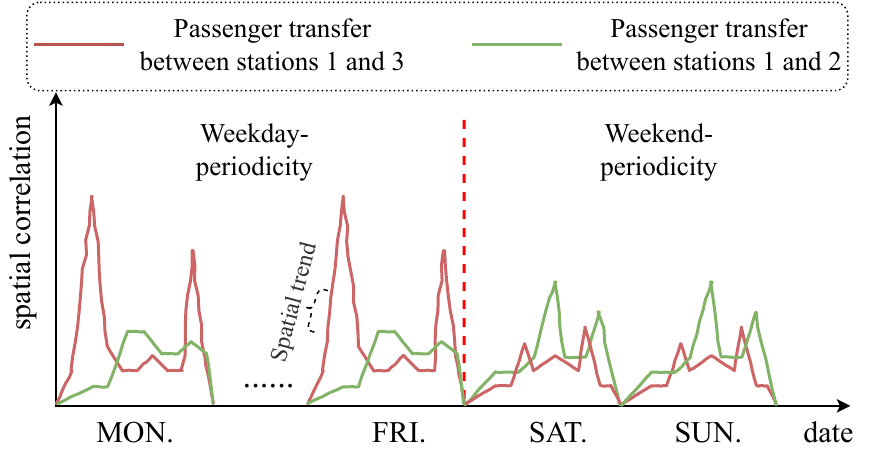}
            \end{minipage}
        \label{fig:motivation2}
        }
	\caption{Spatial correlations with periodicities and trends. The passenger flows between stations in different functional areas show daily spatial trends and distinct weekday and weekend periodicities.}
	\label{fig:motivation}
 \vspace{-10pt}
\end{figure}

Existing spatio-temporal forecasting (STF) methods~\cite{morsali2020spatio,zhang2017deep, ConvLSTM, wu2021autocts,SimpleTS} demonstrate that forecasting accuracy improves significantly when considering both temporal and spatial correlations. For instance, in a metro system, predicting the future outbound passenger flow at one station (e.g., station 1 in Fig.~\ref{fig:motivation1}) requires considering its historical flow (temporal correlation), and the passenger flow at connected stations, e.g., stations 2 and 3 (spatial correlation). The STF problem can be approached as spatio-temporal graph learning, where sensors are treated as graph nodes, and the spatial correlations between sensors are seen as edges. In this framework, time series serve as node features, and a temporal module, such as recurrent neural networks or convolutional neural networks (CNNs), captures temporal correlations for each node. Additionally, graph neural networks (GNNs) are employed to capture hidden spatial dependencies. In addition to enabling sophisticated GNN models, many studies focus on learning optimal graph structures for specific downstream tasks, since the success of GNNs can be attributed to their ability to exploit the potential correlations in graph structures~\cite{zhu2021survey}. In general, graph structures encompass pre-defined graphs~\cite{li2017diffusion,liu2020physical} and  self-learning graphs~\cite{wu2019graph,bai2020adaptive,mtgnn,stemgnn}. The former constructs a task-related graph structure based on domain knowledge, such as geospatial distances between nodes, route topology, or node feature similarity. The latter allows a network to learn a graph structure from data.

Spatial correlations present regular time-varying dynamics, specifically manifested as trends and periodicities. We illustrate these two patterns using a public transportation scenario as an example. As shown in Fig.~\ref{fig:motivation}, the number of Origin-Destination pairs indicates the strength of the spatial correlation between the stations. Specifically, stations 1, 2, and 3 are metro stations located in residential, shopping, and business areas, respectively. \textbf{\textit{Spatial Trend}} denotes an increasing or decreasing correlation over time. The red curve in Fig.~\ref{fig:motivation2} shows a gradual increase during the morning rush hour since people commute to work and then decreases as they reach their destinations. Subsequently, around 18:00, there is an increase in passenger transfers between stations 1 and 3, as well as between stations 2 and 3, as people return home or engage in non-work activities. \textbf{\textit{Spatial Periodicity}} denotes a temporal recurrence of correlations. Passenger transfers between stations show different cyclic daily patterns on weekdays and weekends, as indicated by the dashed red line in Fig.~\ref{fig:motivation2} that separates the two. On weekends, the correlations display significant differences due to the impact of leisure activities and reduced work-related commuting. These dynamics are also observed in air quality, water quality, and other CPS applications, which are affected by the regularity of the underlying processes.

However, existing forecasting solutions are ill-equipped to capture spatial trends and periodicities. Solutions employing pre-defined graphs require excessive computational and storage resources to pre-compute spatial correlations and may also introduce inevitable biases due to incomplete prior assumptions~\cite{liu2020physical}. Solutions employing self-learning graphs either exhibit difficulties in representing dynamics or fail to explicitly consider the regularity of dynamics~\cite{bai2020adaptive}. Overall, three challenges need to be addressed: 
\begin{itemize}
    \item [1.] 
    Learning dynamic graph structures with trends and periodicities has been rarely explored in spatio-temporal forecasting.
    \item [2.]
    Constructing graphs that are capable of representing dynamics, which are time-varying, inevitably introduces more model parameters, making convergence harder.
    \item [3.]
    Dynamic spatio-temporal correlations are difficult to learn. On the one hand, temporal and spatial correlation affect each other dynamically. On the other hand, the current spatio-temporal state is affected by the past state, and this influence propagates and accumulates over time.
\end{itemize}

In this paper, we propose a novel approach, \textit{\underline{T}ime-aware \underline{G}raph \underline{C}onvolutional \underline{R}ecurrent \underline{N}etwork} (TGCRN) framework to tackle the aforementioned challenges. First, we construct graphs with trends and periodicities to represent spatial correlations. Instead of employing sophisticated neural networks, we decompose the graph learning into node and time representations and then blend them to build time-aware graphs. Next, we utilize graph convolution-based gated recurrent units to effectively capture both spatial and temporal dependencies, which combines the graph convolution on time-aware graph structures with the gating mechanism for integrating current input and previous state. Finally, we present TGCRN, which employs an encoder-decoder architecture integrating time-aware graph structure learning and the graph convolution-based gated recurrent unit. This recursive integration allows the model to effectively capture the trends and periodicities of spatio-temporal correlations. Our contributions are four-fold:

\begin{itemize}
    \item Our solution is the first to capture dynamic spatial correlations with trends and periodicities in spatio-temporal forecasting by learning the regular dynamics of graph structures. It opens a new avenue for spatio-temporal analysis research.
    
    \item We propose a novel time-aware graph learning method, incorporating time discrepancy learning and a periodic discriminant function to construct a series of time-aware graphs. Our method enables graph structure learning that adopts a factorized learning perspective, allowing adaptive learning of the dynamics of spatial correlation from data.
        
    \item We develop a holistic model that automatically learns node and time representations and graph structures. Further, it recursively captures regular spatio-temporal dependencies in an end-to-end fashion. This is done by employing an encoder-decoder architecture for multi-step time series forecasting.
    
    \item Experimental results on five real-world datasets show that the proposed method is capable of outperforming the state-of-the-art graph-based approaches. We visualize the learned graph structures, thereby offering insight into the distinct trends and periodicities of spatial correlations over time.
     
\end{itemize}

\begin{figure*}[ht!]
    \centering
    \includegraphics[scale=0.6]{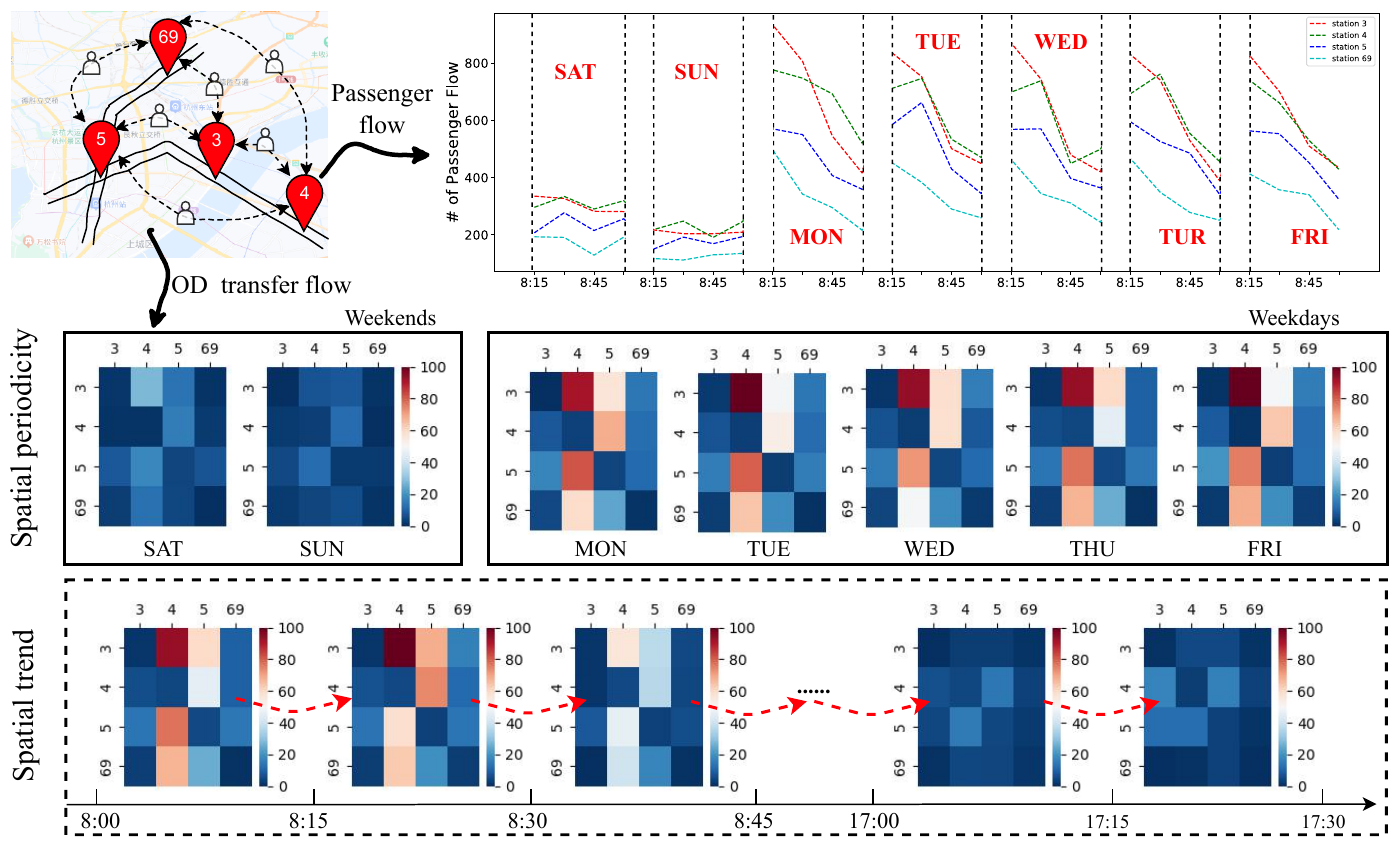}
    \caption{A real-world traffic data example: passenger flows of stations 3, 4, 5, and 69 and the passenger flow transfer between stations from 08:00 to 08:15 on weekdays and weekends, as well as passenger flow transfers over consecutive time spans on one weekday. The amount of passenger transfer represents spatial correlation.}
    \label{fig:pre}
\end{figure*}

\begin{table}[]
\caption{Frequently used notations.}
\label{TAB: NOTATION}
\begin{tabular}{cp{6.5cm}}
\toprule
\textbf{Notation}     & \textbf{Description} \\ \midrule
$N$           & number of time series\\
$P$           & number of recent time steps\\
$d$           & dimensionality of features\\
$Q$           & number of future time steps to forecast  \\
$\mathcal{X}$ & $N$ spatially correlated time series covering $P$ time steps and having d-dimensional features\\
$\mathcal{X}_t$         & $\mathcal{X}$ at timestamp $t$. \\
$\hat{Y}$     & spatially time series of the $Q$ future  time steps \\
$\mathcal{G}$ & graph $\mathcal{G}=(\mathcal{V}, \mathcal{E})$ with node set $\mathcal{V}$, edge set $\mathcal{E}$\\
$\mathcal{A}^t$ & adjacency matrix of graph at timestamp $t$ \\
$E_\nu^i$       & embedding of $i$-th node (time series) \\
$E_\tau^i$      & embedding of $i$-th timestamp\\
$h_t^l$         & hidden state at timestamp $t$ of the $l$-th layer of a model\\
$\mathcal{H}$   & the hidden state at multiple timestamps.\\
$\Phi(\cdot)$   & time encoding function  \\
$\langle \cdot, \cdot \rangle$ & inner product \\ \bottomrule
\end{tabular}
\vspace{-20pt}
\end{table}

\section{Preliminaries}
In this section, we provide related preliminaries on spatio-temporal forecasting. Table~\ref{TAB: NOTATION} summarizes frequently used notation.

\subsection{Definitions}
\begin{mydef*}[Spatially Correlated Time Series] We use $\mathcal{X} = (\mathcal{X}_{t_1}, \mathcal{X}_{t_2}, \cdots, \mathcal{X}_{t_P}) \in \mathbb{R}^{N\times P \times d}$ to denote $N$ spatially correlated multivariate time series, where each time series covers $P$ timestamps with $d$-dimensional features. 
\end{mydef*}

\begin{mydef*}[Graph]
   We use graph $\mathcal{G}=(\mathcal{V}, \mathcal{E})$ to represent spatial correlations between time series, where $\mathcal{V}$ is a set of nodes (representing time series) and $\mathcal{E}$ is a set of weighted edges. An adjacency matrix $\mathcal{A} \in \mathbb{R}^{N \times N}$, where $N=|\mathcal{V}|$, is used to represent the graph. Thus, $\mathcal{A}_{i,j}$ denotes the weight of the edge between nodes $v_i$ and $v_j$, Further, $\mathcal{A}_{i,j} = 0$ means that there is no edge between nodes $v_i$ and $v_j$.
\end{mydef*}

\begin{mydef*}[Time-aware Graph]
    A time-aware graph $\mathcal{G}^t$, which we represent by an adjacency matrix $\mathcal{A}^t$, captures the spatial correlations between correlated time series at time $t$.
\end{mydef*}

\subsection{Spatial Periodicity and Trend}
To verify the spatial patterns with trends and periodicities, we calculate the Origin-Destination (OD) transfer in the Hangzhou metro system. The OD transfers represent the spatial correlations between stations and can be denoted as an adjacency matrix $\mathcal{A}$, where $\mathcal{A}_{i,j}$ denotes the number of passengers from station $i$ to station $j$. As shown in Fig.~\ref{fig:pre}, four stations are located in different areas of Hangzhou. First, we observe that the passenger flows of each station from 08:15 to 09:00 on weekdays are significantly higher than on weekends, and we see that these flows decrease as the morning peak ends. Then we visualize the OD transfer in time interval 08:00 -- 08:15 of the week $\{\mathcal{A}_\text{SAT}^{t_1}, \cdots, \mathcal{A}_\text{FRI}^{t_1}\}$ via heat maps, where timestamp 08:00 is denoted as $t_1$ and where we omit the ending timestamp for brevity. We see that $\mathcal{A}_\text{SAT}^{t_1}$ is similar to $\mathcal{A}_\text{SUN}^{t_1}$ and that $\{\mathcal{A}_\text{MON}^{t_1}, \cdots, \mathcal{A}_\text{FRI}^{t_1}\}$ are similar to each other with minor fluctuation, showing distinct weekend and weekday periodicities because of the demand for work. Moreover, we randomly choose one workday and visualize the spatial correlations over consecutive 15-minute time spans from 08:00 to 17:30, finding a continuous dynamic pattern. For example, the number of passenger transfers from station $5$ to station $4$ decreases gradually from timestamp $t_1$ to $t_4$, i.e., $\mathcal{A}_{5,4}^{t_1} > \mathcal{A}_{5,4}^{t_2} > \mathcal{A}_{5,4}^{t_3} > \mathcal{A}_{5,4}^{t_4}$. 

\subsection{Problem Statement}
In the spatio-temporal forecasting task, given a system of spatially correlated time series, our goal is to learn a function $\mathcal{F}$ that maps historical observations $\mathcal{X}$ to predictions of the following $Q$ future time steps $\hat{Y} = (\hat{y}_{t_{P+1}}, \cdots, \hat{y}_{t_{P+Q}})$. We formulate $\mathcal{F}$ as follows.
\begin{gather}
    (x_{t_1}, x_{t_2}, \cdots, x_{t_P}) \stackrel{\mathcal{F}}{\longrightarrow} (\hat{y}_{t_{P+1}}, \hat{y}_{t_{P+2}}, \cdots, \hat{y}_{t_{P+Q}})
\end{gather}

\section{Methodology}
We proceed to detail the proposed \textit{TGCRN} method. First, we elaborate on how to capture spatial trends and periodicities between time series by enhancing an optimized time-aware graph structure in Section~\ref{section:TagSL}. Then we introduce a \textit{Graph Convolution-based Gated Recurrent Unit} to extract spatio-temporal hidden dependencies in Section \ref{section:GCGRU}. Finally, we present the \textit{Time-aware Graph Convolutional Recurrent Network} framework that integrates \textit{Time-aware graph Structure Learning} and the \textit{Graph Convolution-based Gated Recurrent Unit} with an encoder-decoder architecture for multi-step forecasting in Section~\ref{section:TGCRN}. 

\subsection{Time-aware Graph Structure Learning}\label{section:TagSL}
\subsubsection{Overview of TagSL}
Fig.~\ref{fig:pre} shows that the correlations between time series are dynamic over consecutive time steps, and further exemplify periodicities and trends caused by the underlying processes. To capture such dynamics in graphs, an idea is to pre-construct the graph structure. However, this inevitably yields two problems: 1) high space and time complexity; 2) introducing human bias caused by the priori knowledge-guided metric used to measure correlations between time series, e.g., geographical distance.

\begin{figure}[t!]
    \centering
    \includegraphics[width=\columnwidth]{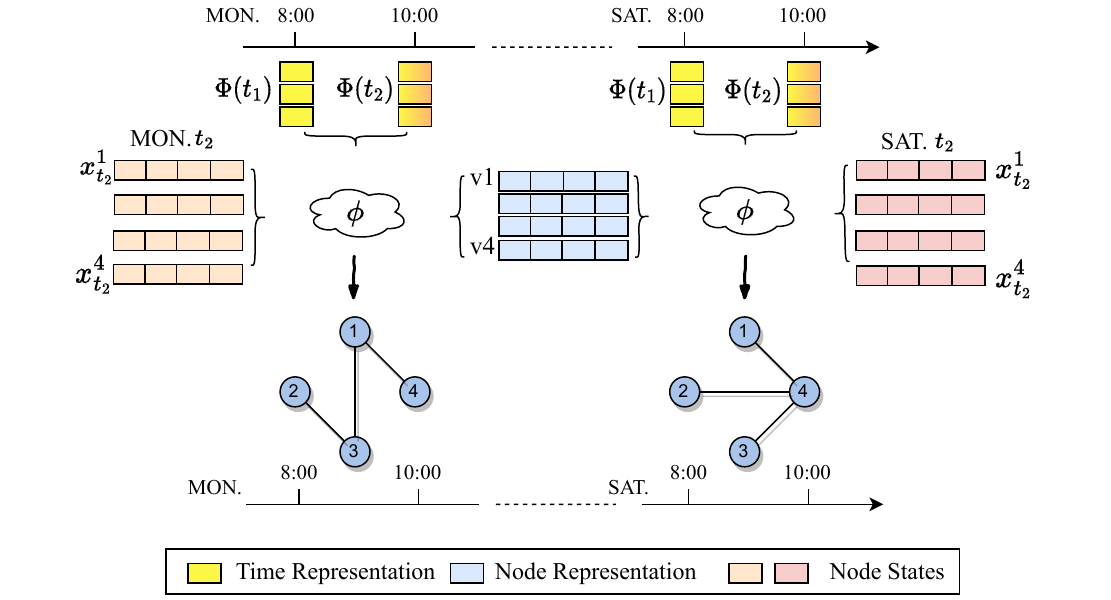}
    \caption{Time-aware graph structure learning.}
    \label{fig:tagsl}
\vspace{-10pt}
\end{figure}

To address these problems, we propose \underline{T}ime-\underline{a}ware \underline{G}raph \underline{S}tructure \underline{L}earning (TagSL), a generic data-driven method. As illustrated in Fig.~\ref{fig:tagsl}, TagSL learns time-aware graph structures by blending of node state and the representations of node and time. Formally, we define $ \phi(E_{\nu}, \Phi(t), \mathcal{X}_{t}):= \mathcal{F}_\mathcal{G}(t)$, where $E_{\nu} \in \mathbb{R}^{N \times d_N}$ denotes the node embedding with $d_N$-dimensional vectors of $N$ nodes, $\Phi(t): t \mapsto \mathbf{R}^{d_T}$ is a time encoding function that maps times to $d_T$-dimensional vectors, $\mathcal{X}^t \in \mathbb{R}^{N\times d}$ is the node state at time step $t$. $\phi(\cdot)$ is a composition function that we study to generate the graph adjacency matrix at a specific time.

\begin{figure}[t]
    \centering
    \includegraphics[width=\columnwidth]{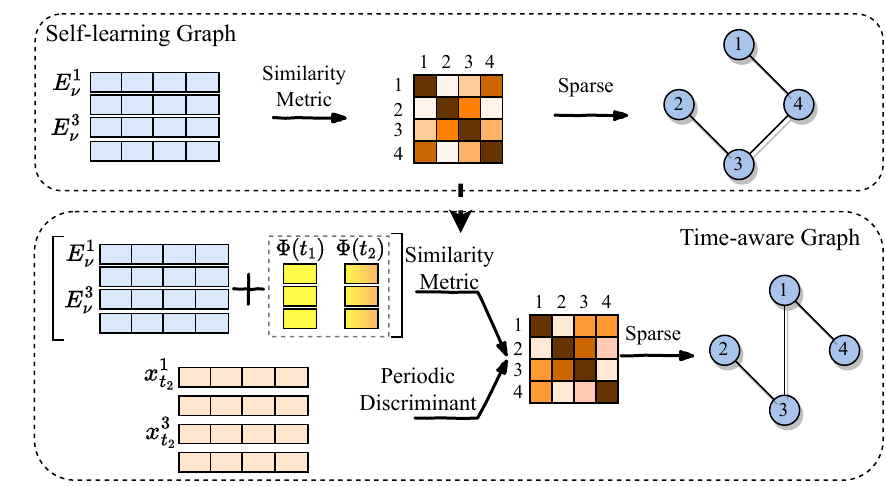}
    \caption{Time-aware graph vs. self-learning graph.}
    \label{fig:tagsl2}
\vspace{-10pt}
\end{figure}

The inspiration for TagSL stems from the self-learning graph \cite{bai2020adaptive, wu2019graph}. As shown in Fig.~\ref{fig:tagsl2}, it uses the inner product of the representations of node pairs i.e., $\langle E_{\nu}^i, E_{\nu}^j \rangle := \mathcal{F}_\mathcal{G}$ to measure edge weights, which can not only learn hidden inter-dependencies between nodes but can also reduce the number of parameters compared to directly learning an adjacency matrix. To build a time-aware graph structure, suppose that the concatenation of the node and time representations, $e_{i,t_1} = [E_{\nu}^i; \Phi(t_1)]$ enable the vector representation of node $i$ to combine $t_1$ time and static node information. Thus, the time-aware correlation between nodes can be defined as follows:
\begin{gather}
\langle e_{i,t_1}, e_{j,t_1} \rangle = \langle E_{\nu}^i,E_{\nu}^j \rangle + \langle \Phi(t_1),\Phi(t_1) \rangle,
\end{gather}
where $\langle \cdot \rangle$ is the inner product operator. Generally, the interactions between time series occur over time, meaning that calculations involving adjacent time steps can reveal additional temporal behavior than calculations at a single time step can. Thus, $\langle \Phi(t_1),\Phi(t_2) \rangle$ would express more meaningful temporal information than $\langle \Phi(t_1),\Phi(t_1) \rangle$. Specifically, $\langle E_{\nu}^i, E_{\nu}^j \rangle$ represents the static spatial correlation between node $i$ and node $j$ and  $\langle \Phi(t_1),\Phi(t_2) \rangle$ intends to represent the temporal evolution of the graph structure. This way, the learning of spatial trend and spatial periodicity is transformed into time representation learning. We further include a time discrepancy learning module to preserve the spatial trend, and a periodic discriminant learning module to distinguish periods.

\begin{algorithm}[t]
    \caption{Time-distance Sampling Algorithm}\label{alg:sampling} 
    \textbf{Input:} Batch discretized time samples $X_\tau \in \mathbb{R}^{B \times T}$. \quad \quad \quad\\
    \textbf{Output:} Anchor samples $X_{\tau_{\mathcal{O}}} \in \mathbb{R}^{B}$, adjacent samples $X_{\tau_{\vartriangle}} \in \mathbb{R}^{B}$, mid-distance samples $X_{\tau_{\lozenge}} \in \mathbb{R}^{B}$, distant samples $X_{\tau_{\triangledown}} \in \mathbb{R}^{B}$.
    
    \begin{algorithmic}[1]
    \Require adjacent range $\gamma_\vartriangle$, mid-distance range $\gamma_\lozenge$, distant range $\gamma_\triangledown$, random choice function $\mathcal{F}:\textit{set} \rightarrow \textit{element}$.
    \State Initialize $X_{\tau_{\mathcal{O}}}=X_{\tau_{\vartriangle}}=X_{\tau_{\lozenge}}=X_{\tau_{\triangledown}}=\varnothing$;
    \State \textbf{for} $i$ in range($B$) \textbf{do}  \Comment{/*Batch loop*/}
    \State \qquad $\tau_i \leftarrow \mathcal{F}(X_\tau^i)$;
    \Comment{/* Randomly choose anchor time from $i$-th row of $X_\tau$*/}
    \State \qquad $X_{\tau_{\mathcal{O}}} \gets \{X_{\tau_{\mathcal{O}}}, X_\tau^{i,\tau_i}\}$;
    \Comment{/* Concatenate anchor sample */}
    \State \qquad $j \leftarrow \mathcal{F}([\tau_i-\gamma_\vartriangle, \tau_i+\gamma_\vartriangle])$;
    \Comment{/* Randomly choose adjacent index based on anchor*/}
    \State \qquad $X_{\tau_{\vartriangle}} \gets \{X_{\tau_{\vartriangle}}, X_\tau^{i,j}\}$; 
    \Comment{/* Concatenate adjacent sample */}
    \State \qquad $k \leftarrow \mathcal{F}([\tau_i-\gamma_\lozenge, \tau_i+\gamma_\lozenge])$;
    \Comment{/* Randomly choose mid-distance index based on anchor*/}
    \State \qquad $X_{\tau_{\lozenge}} \gets \{X_{\tau_{\lozenge}}, X_\tau^{i,k}\}$;
    \Comment{/* Concatenate mid-distance \quad sample */}
    \State \qquad $\tau_j \leftarrow \mathcal{F}([1, B]-{i})$;
    \Comment{/* Randomly choose another row from $X_\tau$*/}
    \State \qquad $d \leftarrow \mathcal{F}(X_\tau^{\tau_j})$;
    \Comment{/* Randomly choose distant index*/}
    \State \qquad $X_{\tau_{\triangledown}} \gets \{X_{\tau_{\triangledown}}, X_\tau^{\tau_j,d}\}$;
    \Comment{/*Concatenate distant sample*/}
    \State \textbf{end for}
    \State \textbf{Return} $X_{\tau_{\mathcal{O}}}$, $X_{\tau_{\vartriangle}}$, $X_{\tau_{\lozenge}}$, $X_{\tau_{\triangledown}}$;
    \end{algorithmic}
\end{algorithm}
\subsubsection{Time Discrepancy Learning}\label{TDL}
The time encoding function $\Phi(\cdot)$ should satisfy two criteria to learn spatial trends. First, it should conform to translation variance, i.e., the metrics of time representations vary over adjacent time steps. Assuming a function $\langle \Phi(t), \Phi(t+c) \rangle := \mathcal{K}(t, t+c) $, translation variance can be formulated as $\mathcal{K}(t, t+c) \ne \mathcal{K}(t+i*c, t+(i+1)*c)$, where $i \ne 0$ and $c$ denotes the time interval. Second, the encoding should preserve discrepancies between time steps. For example, 08:00 and 09:00 should be more similar than 08:00 and 10:00. While several studies, like Time2vec~\cite{kazemi2019time2vec} and TGAT~\cite{xu2019self}, explore model-agnostic and heuristic-driven time representation functions, they primarily focus on the intrinsic properties, such as invariance of time rescaling and the differences in ranges of time steps. 

\begin{figure}[t]
    \centering
    \includegraphics[width=0.7\columnwidth]{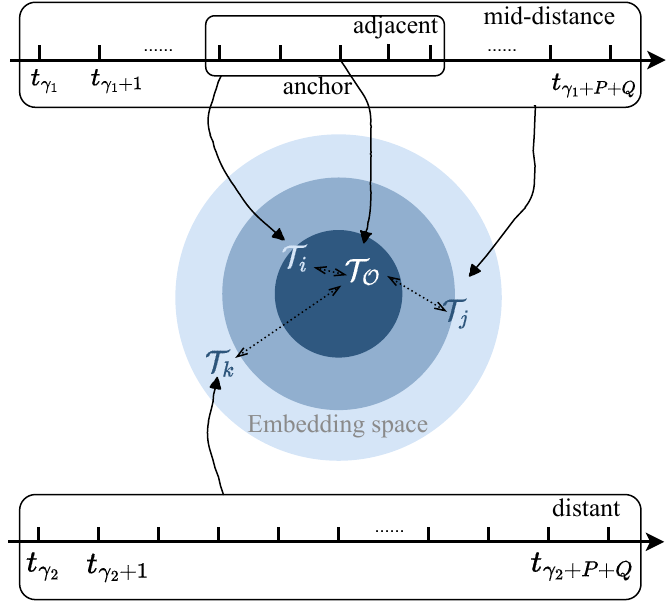}
    \caption{Time discrepancy learning.}
    \label{fig:tagsl3}
    \vspace{-20pt}
\end{figure}

We design a time encoding function based on embedding technology and self-supervised learning. Considering a minimum periodicity such as a day, we first discretize infinite time into a sequence of timestamps of duration a day, denoted as $T=[t_0, t_1, \cdots, t_\text{max}]$. Then we randomly initialize learnable time vectors $E_\tau \in \mathbb{R}^{|T| \times d_T}$ in a finite-dimensional space for all elements, which are optimized using gradient descent. To learn the discrepancies between the vector representations of time steps and enable them to be proportional to the distances in the time domain, we propose a distance-based proportion regularization term to constrain the time embedding. As shown in Fig.~\ref{fig:tagsl3}, there are three different sets of time steps: adjacent, mid-distance, and distant time steps ($|t_{\gamma_1}-t_{\gamma_2}| \gg (P+Q)$). We aim to enable the time representations to be more similar if their specific time steps are closer (e.g., the anchor and an adjacent time step), and the opposite if the two are farther apart (e.g., the mid-distance or distant time steps). This can be achieved by using the following objective loss:
\begin{gather}
    \mathcal{L_{\text{time}}} = \sum_{i,j}||\frac{\zeta_i}{d_i} - \frac{\zeta_j}{d_j}||_1 + \sum_{i,k}||\frac{\zeta_i}{d_i} - \frac{\zeta_k}{d_k}||_1 + \sum_{j,k}||\frac{\zeta_j}{d_j} - \frac{\zeta_k}{d_k}||_1\\
    \text{under with} \quad \zeta_i = \mathcal{F}_\text{sim}(E_\tau^{t_i}, E_\tau^{t_\mathcal{O}})\\
    \text{and} \quad d_i = \mathcal{F}_\text{dist}(t_i, t_\mathcal{O}).
\end{gather}
Here, we omit the expressions of the calculation of $\zeta_j$ and $\quad \zeta_k$, which are the same as $\zeta_i$, except for the time step. $\mathcal{F}_\text{sim}$ denotes the similarity of time representations and $\mathcal{F}_\text{dist}$ denotes the distance between time steps. Considering the similarity measurement in vector space, we utilize the Euclidean distance as $\mathcal{F}_\text{sim}$. To keep the symmetry of distance between time steps, we simplify $\mathcal{F}_\text{dist}$ to $L1$ distance. $t_i$, $t_j$, and $t_k$ denote adjacent, mid-distance, and distant time steps, which all are sampled up to $t_\mathcal{O}$, an anchor time step. The detailed sampling strategy is shown in Algorithm~\ref{alg:sampling}. We randomly select a time step for each sample in a batch as an anchor, and one of the previous or next $\gamma_\vartriangle$ time steps of the anchor is considered as an adjacent one. The one outside the adjacent range in each sample is taken as a mid-distant one, and one of the time steps in other samples is considered as a distant one. Empirically, we set $\gamma_\vartriangle$ half of the length of the input time steps. By involving more general sampled cases, we desire to regularize the model to learn a smooth translation invariance.

\subsubsection{Periodic Discriminant Function}\label{PDF}
To distinguish periods and generate a graph structure that captures corresponding spatial correlations, we design a periodic discriminant function. We observe that the node state is quite distinct at the same daily time in different periods. Taking traffic flow as an example, the traffic flow is different on weekdays and weekends at the same time due to different travel demands, and the pattern can be extracted from the observations to distinguish the two. Hence, we propose a discriminant function that identifies the corresponding period based on the current node state. Specifically, node states can be mapped to different ranges through piecewise nonlinear functions, and the inner product further expands the boundaries.

Formally, combining self-learning graph construction, time representation, and periodic discrimination, we form TagSL. Given the node embeddings $E_{\nu} \in \mathbb{R}^{N \times d_N}$, time representations $E_\tau \in \mathbb{R}^{|T| \times d_T}$, and node features $\mathcal{X} \in \mathbb{R}^{N \times |T| \times d_F}$, the adjacency matrix $\mathcal{A}^t$ of the learned time-aware graph can be formulated as follows.
\begin{gather}
    \mathcal{A}_{\nu} = \langle E_{\nu}, E_{\nu}^T\rangle\\
    \eta_{\tau}^t = \langle E_{\tau}^t, {E_{\tau}^{t-1}}^T\rangle\\
    \mathcal{A}_{\rho} = tanh(\langle \mathcal{X}, \mathcal{X}^T \rangle) \\
    \mathcal{A}^t=(1+\alpha \sigma(\mathcal{A}_p)) \odot (\mathcal{A}_\nu + \eta_\tau^t),
\end{gather}
where $\mathcal{A}_{\nu}, \mathcal{A}_{\rho} \in \mathbb{R}^{N\times N}$ denote the self-learning matrix and periodic discriminant matrix, $\eta_{\tau}$ is a scalar and denotes the trend factor, $\sigma(\cdot)$ is the sigmoid function, $\odot$ denotes the Hadamard product, and $\alpha$ is the saturation factor, a hyperparameter for adjusting the weight of the periodic effect on current spatial correlations. 

\subsubsection{Comparison with existing approaches}
\begin{figure}[t!]
    \centering
    \includegraphics[width=1\columnwidth]{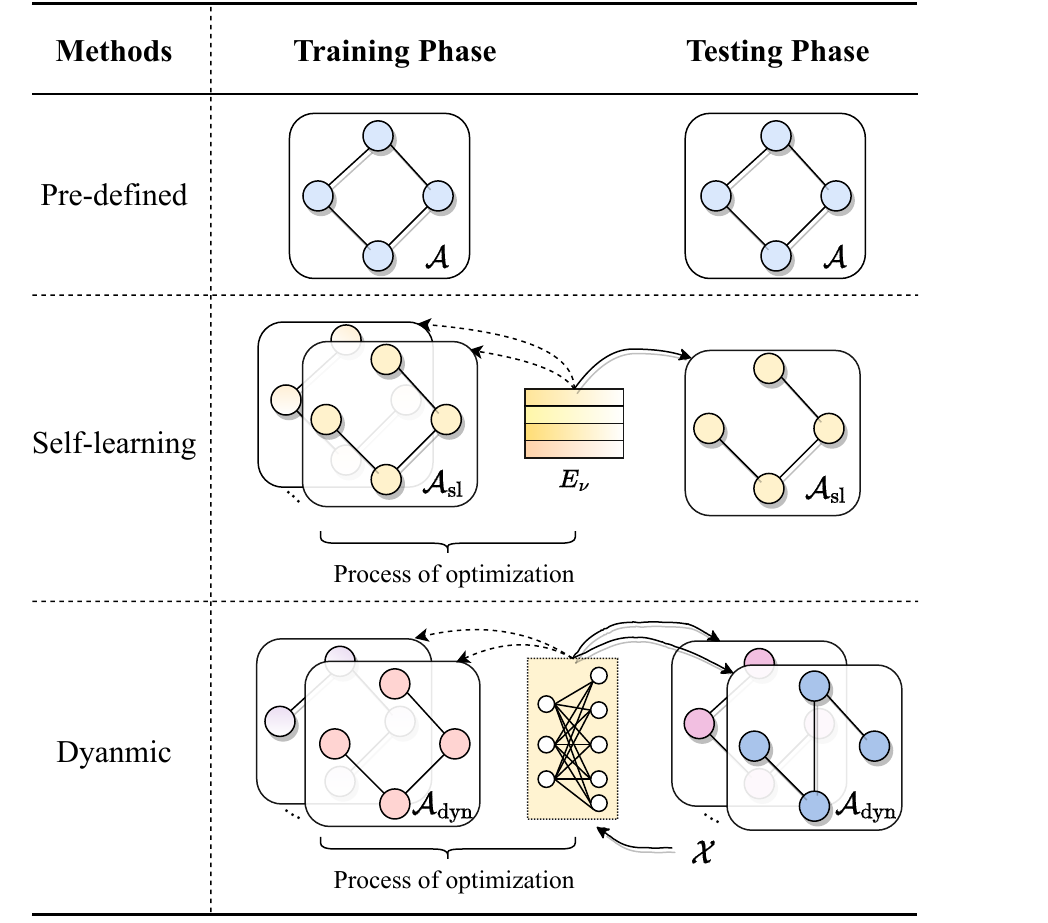}
    \caption{Visual comparison of pre-defined graph, self-learning graph, dynamic graph during training and testing.}
    \label{fig:compare}
\end{figure}

\begin{table}[t!]
\centering
\caption{Comparison with the pre-defined, self-learning, and dynamic graph structures.}
\begin{tabular}{@{}lc@{}}
\toprule
Methods & \multicolumn{1}{l}{Graph Structure Learning} \\ \midrule
Pre-defined~\cite{li2017diffusion,liu2020physical}&  $\mathcal{A}$ \\
Self-learning~\cite{bai2020adaptive,wu2019graph,mtgnn}& $\mathcal{A}_\text{sl}=\phi(E_\nu, E_\nu^T)$ \\
Dynamic~\cite{kim2022graph}  &  $\mathcal{A}_\text{dyn}=\phi(\mathcal{X})$    \\
Time-aware (this paper)  &  $\mathcal{A}^t=\phi(E_\nu, E_\tau^t, \mathcal{X})$ \\ \bottomrule
\end{tabular}
\label{tab:compare}
\vspace{-10pt}
\end{table}

\begin{figure*}[t!]
    \centering
    \includegraphics[width=.9\textwidth]{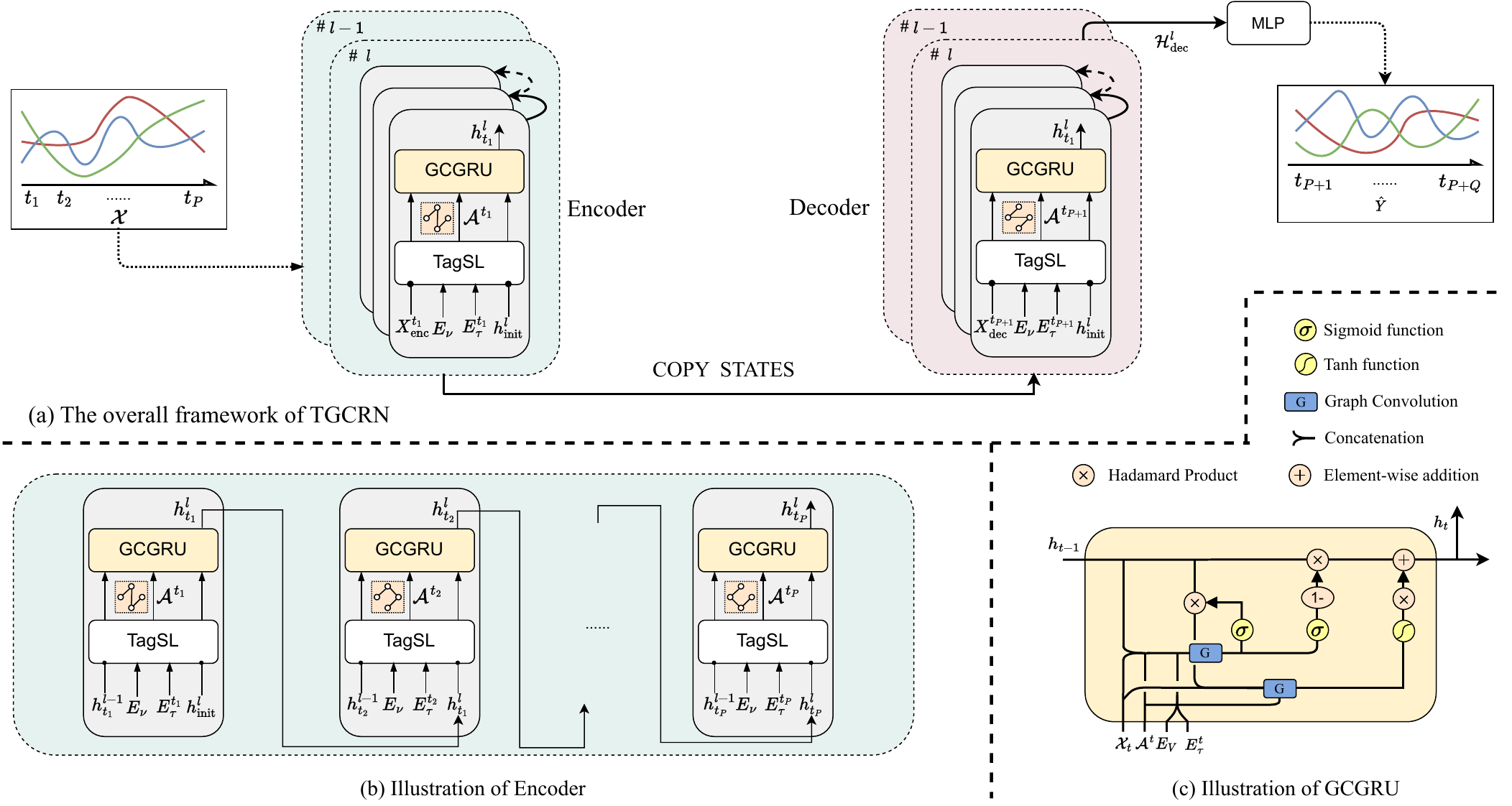}
    \caption{The overall framework of TGCRN}
    \label{fig:framework}
\end{figure*}
First, we visualize how existing methods construct and utilize the graph structures in Fig.~\ref{fig:compare} and give the corresponding formula in Table~\ref{tab:compare}. Generally, a pre-defined graph structure $(\mathcal{A})$ is constructed by domain knowledge and remains fixed during both phases. The self-learning methods derive an optimized graph structure $(\mathcal{A}_\text{sl})$ using a metric function on node embeddings, such as inner product. The pre-defined and self-learning graphs are static for all samples during the testing phase and thus cannot handle dynamic spatial correlations. The dynamic method employs a neural network-based module that uses the nodes' hidden state to generate a series of evolving graph structures $(\mathcal{A}_\text{dyn})$ but lacks an in-depth consideration of regular spatial correlations.

\subsection{Graph Convolution-based Gated Recurrent Unit}\label{section:GCGRU}

Most recent proposals employ graph convolutional networks to capture spatial dependencies between time series with the main objective of learning node representations through message passing. The prominent graph convolutional operation~\cite{GCN} adopts first-order approximations of "Chebyshev polynomial extensions" in the spectral domain. Given the multivariate time series $X \in \mathbb{R}^{N\times C_{in}}$ of $C_{in}$-dimensional feature vectors, the convolution can be expressed as follows.
\begin{gather}
    \begin{gathered}
        Z = L^{\text{sym}}X\mathcal{W}+b,
    \end{gathered}
\end{gather}
where $L^{\text{sym}}$ is a symmetric Laplacian regularization matrix, $\mathcal{W} \in \mathbb{R}^{N \times C_{in} \times C_{out}}$, $b \in \mathbb{R}^{C_{out}}$ are trainable parameters, and $Z \in \mathbb{R}^{N \times C_{out}}$ represents the convolved feature. In addition to capturing the inter-variable correlations, the gated recurrent unit, a variant of recurrent neural networks with a gating mechanism is used for capturing intra-variable temporal patterns. Considering both spatial and temporal dependencies, we propose a graph convolution-based gated recurrent unit that is defined as follows.
\begin{gather}
    \hat{\mathcal{A}^t} = \text{Norm}(\mathcal{A}^t) \\
    \Hat{E}^t = [E_\nu; E_{\tau, t}] \\
    z_t = \sigma(\hat{\mathcal{A}^t}[\mathcal{X}_{:t}; h_{t-1}\Hat{E}^tW_z + \Hat{E}^tb_z)\\
    r_t = \sigma(\hat{\mathcal{A}^t}[\mathcal{X}_{:t}; h_{t-1}]\Hat{E}^tW_r + \Hat{E}^tb_r) \\
    \hat{h}_t = tanh(\hat{\mathcal{A}}^t[\mathcal{X}_{:t}; r_t \odot h_{t-1}]\Hat{E}^tW_{\hat{h}} + \Hat{E}^tb_{\hat{h}})\\
    h_t = (1-z_t) \odot h_{t-1} + z_t \odot \hat{h}_t
\end{gather}
Here, $\hat{\mathcal{A}^t}$ is the adjacency matrix of the time-aware graph at time step $t$, Norm denotes a normalization function, e.g., the softmax function, $tanh$ denotes the hyperbolic tangent function, and $z_t, r_t$, and $\hat{h}_t$ are update, reset, and candidate activation vectors, respectively. Each gate considers the previous hidden state and the current input with learned parameters that include the weight matrix $W \in \mathbb{R}^{C_{in} \times C_{out}}$ and the bias $b \in \mathbb{R}^{C_{out}}$. Meanwhile, to reduce the parameter scale and control overfitting caused by the weight $\mathcal{W} \in \mathbb{R}^{N \times C_{in} \times C_{out}}$, we employ the matrix decomposition $\mathcal{W}=\hat{E}W, E\in \mathbb{R}^{N \times d} \text{ with } d \ll N$, where $E$ is the node representations.

\subsection{Time-aware Graph Convolutional Recurrent Network}\label{section:TGCRN}
Here, we present the overall TGCRN framework, shown in Fig.~\ref{fig:framework}, that adopts an encoder-decoder architecture to output multi-step predictions. To enhance the capacity of feature representation, the encoder and decoder employ a multi-layer network and recursively extract time-aware spatial-temporal correlations. Specifically, given the inputs of $i^{th}$ layer in the encoder or decoder $X_{t_j}^{i}=h_{t_{j-1}}^{i-1}$, the previous hidden state $h^i_{t_{j-1}}$, node embedding $E_\nu$, and time vector $E_{\tau,t_j}$ at time $t_j$, they are first fed to TagSL to obtain a time-aware graph structure $A^{t_j}$. Then the GCGRU is utilized to aggregate the spatial correlation between nodes and their neighborhood derived from $A^{t_j}$, and capture the intra-variables temporal correlation. The output hidden state $h^i_{t_{j}}$ is considered as the input of the next unit. Noting that $X_{t_j, \text{enc}}^{1}=\mathcal{X}_{t_j}$ and $X_{t_j,\text{dec}}^{1}=h_{t_{j}, \text{enc}}^{l}$ when $i=1$. Further, the decoder and the encoder have an identical structure, except for an additional output layer that transforms the hidden state $[h^{l}_{t_{P+1}}, \cdots, h^{l}_{t_{P+Q}}]$ of the last layer in the decoder to an output with the desired dimensionality.

Finally, we present the overall learning objective of TGCRN, including an auxiliary time discrepancy learning loss and error loss. Formally,
 \begin{gather}
    \mathcal{L} = \mathcal{L}_\text{error} + \lambda \mathcal{L}_\text{time}  \\
    \text{with } \mathcal{L}_\text{error} = \frac{1}{|Y|}\sum_i^{|Y|}|Y_i-\hat{Y}_i|,
 \end{gather}
where $\mathcal{L}_\text{error}$ measures the mean absolute error between the ground truth $Y$ and the prediction $\hat{Y}$, $\mathcal{L}_\text{time}$ measures the MAE between timestamps, and $\lambda$ is an adjustable hyperparameter. To sum up, the goal of our task is to optimize all the trainable parameters by minimizing the joint loss objective. 

\begin{table}
\centering
\caption{Summary of datasets.}
    \resizebox{\columnwidth}{!}{
    \begin{tabular}{cccccc} 
        \toprule
        Datasets & City     & Nodes & Time Interval & Length                                    & Partition                    \\ 
        \hline
        \midrule
        HZMetro  & Hangzhou & 80          & 15 mins        & 1825                   & 17d/2d/6d                    \\
        SHMetro  & Shanghai & 288         & 15 mins        & 6716                  & 62d/9d/20d                   \\ 
        \midrule
        NYC-Bike & New York  & 250         & 30 mins        & \multirow{2}{*}{4368} & \multirow{2}{*}{7/1.5/1.5~}  \\
        NYC-Taxi & New York  & 266         & 30 mins        &                                             &                              \\ 
        \midrule
        Electricity & - & 321 & 1 hour & 26304 & \ 7/1/2 \\
        \bottomrule
    \end{tabular}
    }
    \label{tab:dataset}
\end{table}

\begin{table*}
\centering
\caption{Overall forecasting performance on the SHMetro and HZMetro datasets. methods with '*' indicate that the findings come from a previous study, and '$\_$' is used to highlight the best-performing baseline.}
\label{Tab:metro}
\resizebox{2\columnwidth}{!}{
\begin{tabular}{ccccclccclccclccc} 
\toprule
\multirow{2}{*}{Dataset} & \multirow{2}{*}{Method} & \multicolumn{3}{c}{15 min} &  & \multicolumn{3}{c}{30 min} &  & \multicolumn{3}{c}{45 min} &  & \multicolumn{3}{c}{60 min}  \\ 
\cline{3-5}\cline{7-9}\cline{11-13}\cline{15-17}
                         &                         & MAE   & RMSE   & MAPE\%      &  & MAE   & RMSE   & MAPE\%      &  & MAE   & RMSE   & MAPE\%      &  & MAE   & RMSE   & MAPE\%       \\ 
\hline\hline
\multirow{9}{*}{HZMetro} & HA                      & 51.43          & 111.86         & 25.31          &  & 51.38          & 111.80         & 25.3           &  & 51.11          & 111.64         & 25.36          &  & 50.62          & 111.3          & 25.50           \\
& GBDT         & 36.31          & 57.49          & 19.51          &  & 39.17          & 58.76          & 20.50          &  & 42.78          & 60.27          & 20.84          &  & 47.35          & 64.14          & 22.05\\
& FC-LSTM      & 26.85          & 48.27          & 18.90          &  & 27.45          & 49.59          & 19.35          &  & 28.14          & 51.49          & 20.17          &  & 30.34          & 53.68          & 21.30\\
& Informer      &31.97           &59.22           & 34.34          &  &31.98           & 59.55          & 31.14          &  & 34.45          & 63.65          & 34.25          &  & 38.35          & 70.53          & 40.54\\
& Crossformer  & 28.34          & 51.39          & 36.14          &  & 31.68          & 57.43          & 39.43          &  & 34.65          & 62.71          & 42.31          &  & 38.53          & 69.69          & 44.97\\

& DCRNN                   & 23.93          & 40.78          & 14.79          &  & 24.86          & \underline{42.24}          & 15.43          &  & 25.64          & \underline{43.45}          & 16.40          &  & \underline{26.78}  & \underline{45.42}  & 17.70\\
& Graph WaveNet           & 25.38          & 43.15          & 17.44          &  & 26.61          & 45.24          & 16.87          &  & 27.47          & 48.92          & 18.62          &  & 29.87          & 51.74          & 22.52\\
& AGCRN                   & 24.02          & 42.19          & \underline{14.73}          &  & 25.21          & 44.46          & 15.50          &  & 26.48          & 47.06          & 16.79          &  & 27.53          & 48.48          & 19.74\\
& PVCGN                   & 23.96  & \underline{40.72}  & 14.77  &  & 25.18          & 42.97          & \underline{15.37}  &  & \underline{25.41}  & 44.91          & 16.30          &  & 27.17          & 47.18          & \underline{17.68} \\
& ESG                     & \underline{23.86}          & 41.00          & 14.75          &  & \underline{24.72}  & 42.36  & 15.58          &  & 25.81          & 44.45  & \underline{15.78}  &  & 27.38          & 47.05          & 17.93\\ 
\cmidrule{2-17}
& \textbf{TGCRN}        & \textbf{21.73} & \textbf{35.91} & \textbf{13.65} &  & \textbf{22.33} & \textbf{36.88} & \textbf{13.96} &  & \textbf{23.13} & \textbf{38.40} & \textbf{14.69} &  & \textbf{23.85} & \textbf{39.92} & \textbf{15.87}\\ 
\hline\hline
\multirow{9}{*}{SHMetro} & HA                      & 48.26 & 136.97 & 31.55     &  & 47.88 & 136.81 & 31.49     &  & 47.26 & 136.45 & 31.27     &  & 46.4  & 135.72 & 30.80       \\
& GBDT                    & 32.72 & 62.59  & 23.40     &  & 39.50 & 82.32  & 28.17     &  & 49.14 & 113.95 & 40.76     &  & 57.31 & 137.50 & 52.60      \\
& FC-LSTM                 & 26.68 & 55.53  & 18.76     &  & 27.25 & 57.37  & 19.04     &  & 28.08 & 60.45  & 19.61     &  & 28.94 & 63.41  & 20.59      \\
& Informer      &31.44           &62.01           & 33.26          &  &32.02           & 63.36          & 32.96          &  & 33.81          & 67.08          & 35.55          &  & 37.19          & 71.64          & 40.54\\
& Crossformer  & 32.93          & 63.54          & 47.08          &  & 33.84          & 68.49          & 44.28          &  & 38.61          & 79.09          & 51.98          &  & 40.36          & 84.99          & 49.30\\
& DCRNN                   & 24.04 & 46.02  & 17.82     &  & 25.23 & 49.90   & 18.35     &  & 26.76 & 54.92  & 19.3      &  & 28.01 & 58.83  & 20.44      \\
& Graph WaveNet           & 24.91 & 46.98  & 20.05     &  & 26.53 & 51.64  & 20.38     &  & 28.78 & 58.50  & 21.99     &  & 30.9  & 65.08  & 24.36      \\
& AGCRN                   & 24.50 & 50.01  & 18.37     &  & 25.28 & 52.38  & 19.96     &  & 26.62 & 56.74  & 20.71     &  & 27.5  & 60.45  & 22.46      \\
& PVCGN*                  & \underline{23.29}  & \underline{44.97}  & \underline{16.83}  &  & \underline{24.16}  & \underline{47.83}  & \underline{17.23}  &  & \underline{25.33}  & \underline{52.02}  & \underline{17.92}  &  & \underline{26.29}  & \underline{55.27}  & \underline{18.69} \\
& ESG                     & 25.74          & 49.24          & 19.44          &  & 26.68          & 52.23          & 19.83          &  & 27.67          & 55.72          & 21.45          &  & 28.70          & 58.71          & 22.99 \\
\cmidrule{2-17}
& \textbf{TGCRN}        & \textbf{21.81} & \textbf{43.20} & \textbf{15.87} &  & \textbf{22.51} & \textbf{45.54} & \textbf{16.17} &  & \textbf{23.04} & \textbf{47.56} & \textbf{16.60} &  & \textbf{23.34} & \textbf{48.89} & \textbf{17.06}\\
\bottomrule
\end{tabular}
}
\end{table*}

\section{Experiments}
We proceed to report on comprehensive experiments on five large real-world datasets to answer four questions:

Q1. How does TGCRN perform at spatio-temporal forecasting compared to competing approaches, especially graph-based methods? (Section~\ref{EXP:ALL})

Q2. What are the impacts of the different components in TGCRN? (Section~\ref{EXP:AS})

Q3. Do the learned time-aware graphs align with spatial trends and periodicities? (Section~\ref{EXP:TagSL1})

Q4. Does the learned time representation satisfy to the desired sequence constraint? (Section~\ref{EXP:TagSL2})

\subsection{Experimental Setup}

\subsubsection{Datasets}
The experiments are conducted on five real-world datasets: HZMetro and SHMetro \cite{liu2020physical}, as well as NYC-Bike and NYC-Taxi \cite{ye2021coupled}, and Electricity~\cite{informer,crossformer}. The former two are collected from the metro systems of Hangzhou and Shanghai, China. HZMetro contains $58.75$ million transaction records from $80$ stations from Jan./01/2019 to Jan./25/2019. SHMetro contains $811.44$ million transaction records from $288$ stations from Jul./01/2016 to Sept./30/2016. Each record contains the passenger ID, entry or exit station, and the corresponding timestamp. For each station, the inflow and outflow every 15 minutes are measured by counting the number of passengers who enter or exit the station. The historical flow length $P$ is set to 4 time steps (1 hour), and we predict the values of the inflow and outflow of all stations for the next 4 time steps (1 hour).

The NYC-Bike dataset\footnote{https://github.com/Essaim/CGCDemandPrediction} contains bike sharing records of people's daily usages in New York City, and each record contains a pick-up dock, a drop-off dock, and the corresponding timestamps. Each dock is considered a station, yielding 250 stations in total. The NYC-Taxi dataset\footnote{https://www1.nyc.gov/site/tlc/about/tlc-trip-record-data.page} is collected from NYC OpenData and consists of 35 million taxicab trip records, where each record contains the pick-up longitude and latitude, the drop-off longitude and latitude, and the corresponding timestamps. NYC-Taxi is dockless-based, and 266 virtual stations are formed by clustering the records. The pick-up and drop-off demands for both datasets are measured every 30 minutes. Both datasets range from Apr./01/2016 to Jun./30/2016. The historical length $P$ is 12 time steps (6 hours), and the prediction length $Q$ is 12 time steps (6 hours).

The Electricity dataset\footnote{https://archive.ics.uci.edu/dataset/321/electricityloaddiagrams20112014} records electricity consumption in kWh every 1 hour from 2012 to 2014. $P$ and $Q$ are both set to 12 time steps (12 hours). Brief statistics of five datasets are given in Table~\ref{tab:dataset}.

We process the dataset as in previous studies~\cite{liu2020physical, ye2021coupled}, except that we re-split HZMetro into a training set (Jan.~1--Jan.~19), a validation set (Jan.~20--Jan.~21), and a testing set (Jan.~22--Jan.~25). Traffic patterns are different during the original validation and testing periods, where the original validation set (Jan.~19--Jan.~20, i.e., Saturday and Sunday), and testing set (Jan.~21--Jan.~25, i.e., workdays), fail to verify the general effectiveness of the forecasting method.

\subsubsection{Methods}
We compare TGCRN with thirteen existing time series forecasting methods, including the latest spatio-temporal forecasting methods and transformer-based methods:
\begin{itemize}
    \item Historical Average (HA), a statistical approach calculating the average of the corresponding historical periods as the forecast values.
    \item GBDT~\cite{friedman2001greedy}, a weighted ensemble model consisting of a set of weak learners employing the gradient descent boosting paradigm.
    \item XGBoost~\cite{chen2016xgboost}, a scalable tree-boosting system for both regression and classification tasks.
    \item LSTM~\cite{sutskever2014sequence}, a recurrent neural network variant with gating mechanisms.
    \item Informer~\cite{informer} and Crossformer~\cite{crossformer} employ the transformer architecture for long-term and multivariate time series forecasting, respectively.
    \item DCRNN~\cite{li2017diffusion}, an encoder-decoder architecture with gated recurrent units and diffusion convolution on a pre-defined distance-based graph structure for learning spatio-temporal dependencies.
    \item Graph WaveNet~\cite{wu2019graph} employs graph convolution on a self-learning adjacency matrix and stacked dilated 1D convolution to capture spatial and temporal correlations, respectively.  
    \item AGCRN~\cite{bai2020adaptive}, adaptive graph convolutional recurrent network, performs graph convolution on a self-learning graph and employs gated recurrent units to model inter-dependencies among nodes and intra-node temporal correlations. 
    \item PVCGN~\cite{liu2020physical}, physical-virtual collaboration graph network, integrates multiple pre-defined graphs into graph convolution gated recurrent units for learning spatio-temporal representations.
    \item CCRNN~\cite{ye2021coupled} adopts different self-learning graphs in different layers of GCNs and provides a layer-wise coupling mechanism to bridge the adjacency matrices of the upper and lower levels.
    \item GTS~\cite{gts} combines a discrete graph structural learner and recurrent neural network for spatial and temporal forecasting.
    \item ESG~\cite{ye2022learning} learns a multi-scale dynamic graph through gated recurrent units and combines graph convolution and dilated convolution to capture evolving spatio-temporal representations.
\end{itemize}

\subsubsection{Evaluation Metrics}
For consistency of performance evaluation with the previous studies~\cite{liu2020physical, ye2022learning, informer}, we use Mean Squared Error (MSE), Mean Absolute Error (MAE), and Root Mean Squared Error (RMSE) as the common evaluation metrics. Pearson Correlation Coefficient (PCC) is used to measure the linear correlation for traffic demand forecasting, and Mean Absolute Percentage Error (MAPE) is used to evaluate the relative error for traffic flow forecasting. For PCC, a higher value indicates better performance; for the others, the opposite holds.

\subsubsection{Implementation Details}
We implemented all experiments on an Intel(R) Xeon(R) Gold 5215 CPU \@@ 2.50GHz and two Nvidia Quadro RTX 8000 GPUs. Following \citet{bai2020adaptive}, we adopt the Adam \cite{kingma2014adam} optimizer to update model weights. The L2 penalty is $10^{-4}$. The initial learning rate is $10^{-3}$ and decays by $0.3$ when the number of epochs reaches $[5,20,40,70,90]$. For all datasets, the batch size is $16$. The saturate factor of the periodic discriminant function is set to $0.3$. The numbers of layers of the encoder and decoder and the hidden units of the GCGRU are set to $2$ and $64$, respectively. For HZMetro, we use a node embedding dimensionality of $64$ and a time embedding dimensionality of $32$. For the other datasets, both embeddings have a dimensionality of $64$. We use the early stopping strategy to select the best model weights when the patience reaches $15$.

\subsection{Main Results}\label{EXP:ALL}
\begin{figure}
	\centering
	\subfigure[MAE on HZMetro]{
		\begin{minipage}[b]{0.8\columnwidth}
                \centering
			\includegraphics[width=0.8\columnwidth]{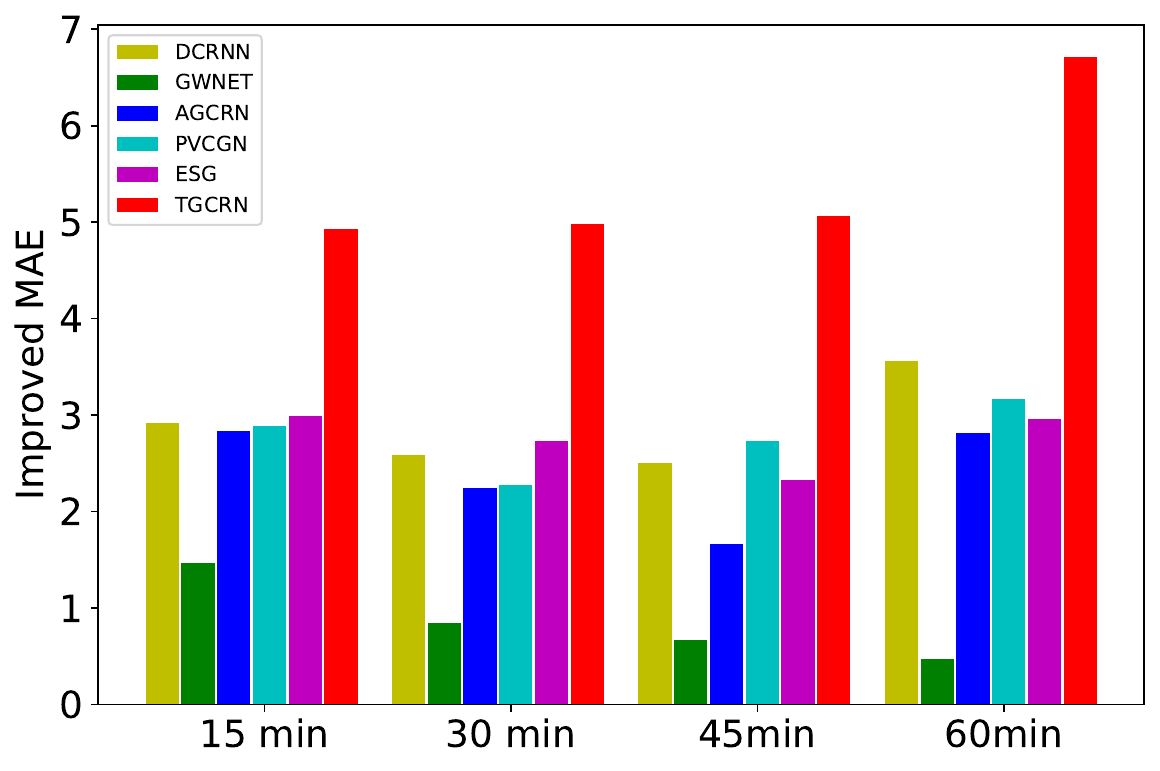}
		\end{minipage}
		\label{fig:mae}
	}
    	\subfigure[RMSE on HZMetro]{
    		\begin{minipage}[b]{0.8\columnwidth}
                \centering
   		 	\includegraphics[width=0.8\columnwidth]{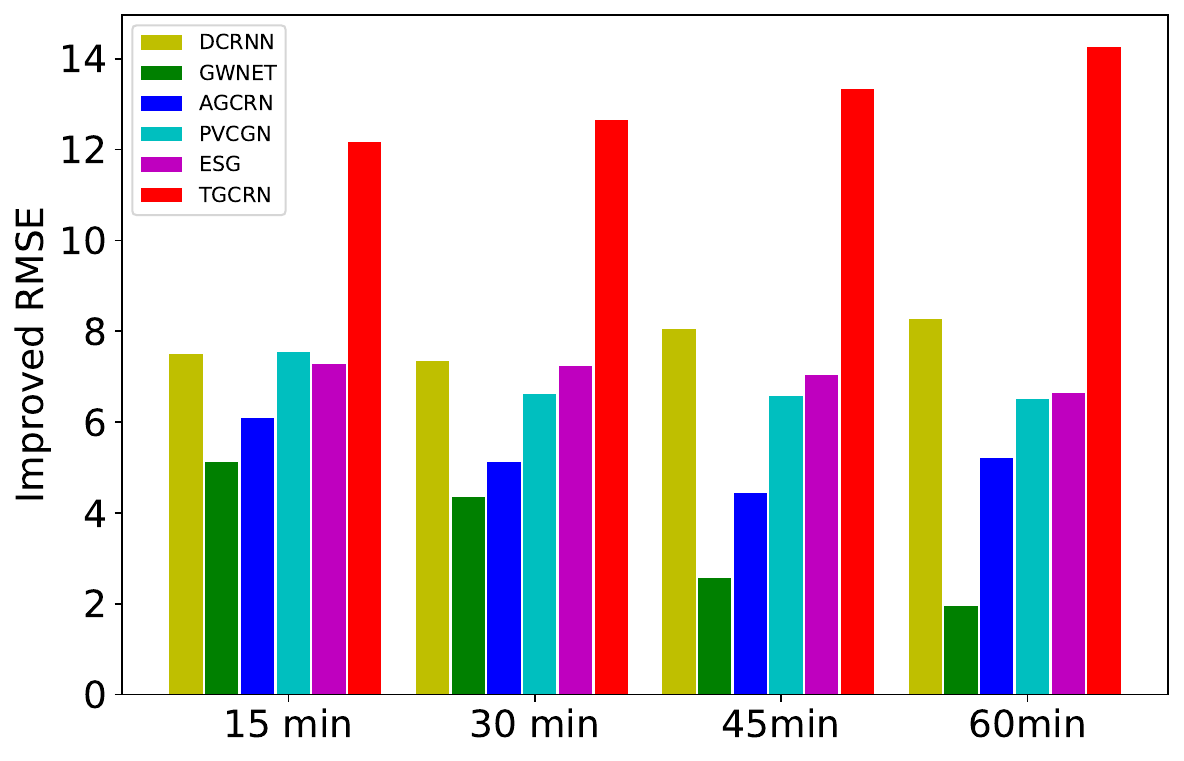}
    		\end{minipage}
		\label{fig:rmse}
    	}
	\caption{Comparison on multiple time steps against FC-LSTM benchmark.}
	\label{fig:multiple}
\end{figure}

Tables~\ref{Tab:metro}, \ref{tab:nyc} and \ref{tab:elc} present the forecasting performance on the five datasets. The results of the baselines on SHMetro, NYC-Bike, and NYC-Taxi are taken from previous studies \cite{liu2020physical,ye2021coupled,ye2022learning}. For HZMetro, due to the reorganization of the dataset, we produce the results using official source codes from the corresponding papers. Overall, we can find that 1) our method consistently achieves the best performance at all metrics, both for all datasets and forecasting horizons (\textbf{Q1}); 2) the methods that capture both spatial and temporal relationships show significant improvement over those capturing only temporal correlations (such as HA, GBDT, and FC-LSTM); 3) from the perspective of spatial correlation modeling, TGCRN outperforms the existing GCN-based approaches that adopt fixed, self-learning and dynamic graph structures, including DCRNN, AGCRN, and ESG, demonstrating the effectiveness of capturing time-aware spatial correlations. More specifically, TGCRN achieves $10.95\%$ and $14.16\%$ improvements on HZMetro, $8.44\%$ and $7.44\%$ improvements on SHMetro, and $6.15\%$ and $6.33\%$ improvements on NYC-Taxi in terms of MAE and RMSE with average horizons, Compared with existing state-of-the-art methods. TGCRN outperforms ESG due to its enhanced capacity to capture regular dynamics of spatial correlations; 4) from the perspective of multi-step forecasting, TGCRN consistently maintains superiority. As shown in Fig.~\ref{fig:multiple}, taking FC-LSTM as a benchmark, we find that as the time step increases, TGCRN shows more and more prominent predictive performance compared to the other methods. We further observe that ESG and Graph WaveNet struggle to extract meaningful temporal dependencies with their CNN-based temporal modules, limited by the short-term setting ($P=4, Q=4$) on the metro datasets, which widens the gap to TGCRN on NYC-Bike and NYC-Taxi. We also note that although PVCGN utilizes pre-defined graph structures, it achieves top-tier performance, benefiting from the ability of multiple graph learning. Yet, it requires more hand-crafted engineering and incurs higher computing costs (more discussion in Section~\ref{sec:cost}).

\begin{table}
\centering
\caption{Overall forecasting performance on the NYC Bike and NYC- Taxi datasets. The cells with '-' indicate that the corresponding metric was not reported in the original paper.}
\label{tab:nyc}
\resizebox{\columnwidth}{!}{
\begin{tabular}{cccclccc} 
\toprule
\multirow{2}{*}{Method} & \multicolumn{3}{c}{NYC Bike}                        &  & \multicolumn{3}{c}{NYC Taxi}                         \\ 
\cline{2-4}\cline{6-8}
& MAE             & RMSE            & PCC             &  & MAE             & RMSE            & PCC              \\ 
\hline\hline
HA                      & 3.4617          & 5.2003          & 0.1669          &  & 16.1509         & 29.7806         & 0.6339           \\
XGBoost                 & 2.4689          & 4.0494          & 0.4107          &  & 11.6806         & 21.1994         & 0.8077           \\
FC-LSTM                 & 2.3026          & 3.8139          & 0.4861          &  & 10.2200         & 18.0708         & 0.8645           \\
Informer                 & 1.7650          & 2.8341          & -          &  & 5.7888         & 18.0708         & -           \\
Crossformer                 & 2.0908          & 3.2898          & -          &  & 5.9777         & 10.5976         & -           \\
DCRNN                   & 1.8954          & 3.2094          & 0.7227          &  & 8.4274          & 14.7926         & 0.9122           \\
Graph WaveNet           & 1.9911          & 3.2943          & 0.7003          &  & 8.1037          & 13.0729         & 0.9322           \\
CCRNN*                  & 1.7404          & 2.8382          & 0.7934          &  & 5.4979          & 9.5631          & 0.9648           \\
GTS                     & 1.7798          & 2.9258          & -               &  & 7.2095          & 12.7511         & -                \\
ESG*                    & \underline{1.6129}          & \underline{2.6727}          & -               &  & \underline{5.0344}  & \underline{8.9759}  & -                \\ 
\midrule
\textbf{TGCRN}        & \textbf{1.5889} & \textbf{2.6106} & \textbf{0.8319} &  & \textbf{4.7244} & \textbf{8.4074} & \textbf{0.9725}  \\
\bottomrule
\end{tabular}
}
\end{table}

\begin{table}[]
\centering
\caption{Forecasting performance on the Electricity dataset.}
\begin{tabular}{ccc}
\hline
\multirow{2}{*}{Method} & \multicolumn{2}{c}{Electricity} \\ \cline{2-3} 
                        & MSE            & MAE            \\
\hline\hline
Graph WaveNet           & 0.2313              & 0.3226              \\
AGCRN                   & 0.1725              & 0.2756              \\
Informer                & 0.2330         & 0.3453         \\
Crossformer             & 0.1453         & 0.2620              \\
ESG                     & 0.1563              & 0.2651              \\ \hline
TGCRN                   & \textbf{0.1440}         & \textbf{0.2517}         \\ \hline
\end{tabular}
\label{tab:elc}
\vspace{-20pt}
\end{table}

\subsection{Model Analysis}
\subsubsection{Ablation Study}\label{EXP:AS}
We conduct ablation studies to understand the impact of the time-aware graph structure learning and encoder-decoder architecture (\textbf{Q2}). First, we design four variants of our graph learning mechanism.
\begin{itemize}
    \item \textit{w/o tagsl} replaces the time-aware graph structure learning with the self-learning mechanism of AGCRN.
    \item \textit{w/ TE} only utilizes time embedding in the time-aware graph structure learning. 
    \item \textit{w/o TDL} removes the time discrepancy learning to assess the effect of the learned time representation.
    \item \textit{w/o PDF} removes the periodic discriminant function to assess its contribution by capturing the effects of different periodicities.
\end{itemize}
Second, we utilize the most recent prominent time representation methods: Time2vec~\cite{kazemi2019time2vec} and continuous-time representation~\cite{xu2019self} for encoding time to assess the effect of our simple but effective time representation. The variants \textit{Time2vec} and \textit{CTR} replace our time embedding and time discrepancy learning. Finally, \textit{w/o enc-dec} replaces the recursively obtained decoding output with direct output based on a fully connected neural network.

Table \ref{tab:as} reports the results for variants of TGCRN on HZMetro and SHMetro. First, we observe that \textit{w/o tagsl} suffers a significant drop in performance, which indicates that time-aware graph structures provide more accurate dynamics of spatial correlations. Next, the results for \textit{w/ TE} indicate that the discretized time embedding used for making the time-aware graph is important for spatial patterns. However, leveraging the time embedding alone cannot guarantee the learning of a meaningful time representation, since the model trivially optimizes the representation based on the forecast loss of a downstream task, as mentioned in Section \ref{section:TagSL}. The results for \textit{w/o TDL} and \textit{w/o PDF} show that time discrepancy learning and the periodic discriminant function are both crucial for spatio-temporal forecasting. In the further validation of our time representation, the results for \textit{Time2vec} and \textit{CTR} show that the combination of time embedding and time discrepancy learning is more suitable for our model. The result of \textit{w/o enc-dec} suggests that iteratively predicting future values over multiple time steps helps the model to better capture spatio-temporal dependencies.

\begin{table}[]
\caption{Ablation study on HZMetro and SHMetro. TE, TDL, and PDF denote Time Embedding, Time Discrepancy Learning, and Periodic Discriminant Function, respectively.}
\begin{tabular}{lcccccc}
\toprule
\multirow{2}{*}{Methods}       & \multicolumn{3}{c}{HZMetro} & \multicolumn{3}{c}{SHMetro}  \\ \cline{2-7} 
& MAE     & RMSE    & MAPE    & MAE   & RMSE    & MAPE     \\ 
\hline\hline
TGCRN      & \textbf{22.71}   & \textbf{37.76}   & \textbf{14.54}   & \textbf{22.68}  & \textbf{46.30}      & \textbf{16.43}\\ 
\midrule
\textit{w/o tagsl}  & 25.40   & 44.52   & 15.85   & \multicolumn{1}{l}{26.99} & \multicolumn{1}{l}{57.10} & \multicolumn{1}{l}{20.07} \\
\textit{w/ TE}  & 22.90   & 38.05   & 14.74   & 23.36 & 46.83  & 17.43  \\
\textit{w/o TDL} & 22.84   & 38.02   & 14.89   & \multicolumn{1}{l}{22.85} & \multicolumn{1}{l}{46.32} & \multicolumn{1}{l}{16.76} \\
\textit{w/o PDF} & 22.78   & 37.69   & 14.70   & \multicolumn{1}{l}{23.26} & \multicolumn{1}{l}{46.74} & \multicolumn{1}{l}{17.33} \\ 
\midrule
\textit{Time2vec}\cite{kazemi2019time2vec} & 25.95   & 47.94   & 15.77   &25.14   &61.90   &17.57   \\
\textit{CTR}\cite{xu2019self}  & 23.16   & 39.51   & 14.73   &23.81   &49.36   &16.96   \\ 
\midrule
\textit{w/o enc-dec} & 22.91 & 38.23   & 14.59   & 24.35  & 51.47  & \multicolumn{1}{l}{18.22} \\ \bottomrule
\end{tabular}
\label{tab:as}
\vspace{-20pt}
\end{table}

\begin{figure}[!t]
    \centering
    \subfigure[]{
        \centering
        \begin{minipage}[b]{0.5\columnwidth}
            \resizebox{\linewidth}{!}{
            \begin{tikzpicture}[scale=0.6]
                \begin{axis}[
                    xlabel={size of $d_\tau$},
                    ylabel={MAE}, 
                    label style={font=\LARGE},
                    tick label style = {font=\Large},
                    xticklabels={16,32,64,128}, 
                    xtick={1,2,3,4},
                    ymin=22.50, ymax=24.50,
                    legend entries = {$d_\nu=16$, $d_\nu=32$, $d_\nu=64$},
                    legend columns = -1,
                    legend style={nodes={scale=0.55, transform shape}, font=\huge},
                    legend to name={legendperformance1},
                    legend pos= north west,			
                    ymajorgrids=true,
                    grid style=dashed,
                    y tick label style={/pgf/number format/.cd,%
                        scaled y ticks = false,
                        set thousands separator={},
                        fixed
                    },
                    ]
                    \addplot[
                    color=blue,
                    mark=square,
                    mark size=4pt,
                    ]
                    coordinates {
                        (1,24.35)(2,24.02)(3,23.79)(4,23.34)
                    };
                    \addplot[
                    color=green,
                    mark=asterisk,
                    mark size=4pt,
                    ]
                    coordinates {
                        (1,23.84)(2,23.21)(3,23.08)(4,22.87)
                    };
                    \addplot[
                    color=red,
                    mark=diamond,
                    mark size=4pt,
                    ]
                    coordinates {
                        (1,23.24)(2,22.72)(3,22.77)(4,22.66)
                    };
                \end{axis}
            \end{tikzpicture}
        }
        \end{minipage}
        \label{subfigure1}	
    }%
    \subfigure[]{
        \centering
        \begin{minipage}[b]{0.5\columnwidth}
		\resizebox{\linewidth}{!}{
			\begin{tikzpicture}[scale=0.6]
				\begin{axis}[
					xlabel={size of $d_\tau$},
					ylabel={RMSE}, 
					label style={font=\LARGE},
					tick label style = {font=\Large},
					xticklabels={16,32,64,128}, 
					xtick={1,2,3,4},
					ymin=37.00, ymax=42.30,
					legend entries = {$d_\nu=16$, $d_\nu=32$, $d_\nu=64$},
                        legend columns = -1,   
					legend style={nodes={scale=0.55, transform shape},font=\huge},
					legend to name={legendperformance1},
					legend pos= north west,			
					ymajorgrids=true,
					grid style=dashed,
					y tick label style={/pgf/number format/.cd,%
						scaled y ticks = false,
						set thousands separator={},
						fixed
					},
					]
					\addplot[
					color=blue,
					mark=square,
					mark size=4pt,
					]
					coordinates {
						(1,42.08)(2,40.71)(3,40.16)(4,39.01)
					};
					\addplot[
					color=green,
					mark=asterisk,
					mark size=4pt,
					]
					coordinates {
						(1,41.21)(2,38.89)(3,38.68)(4,37.98)
					};
					\addplot[
					color=red,
					mark=diamond,
					mark size=4pt,
					]
					coordinates {
						(1,39.60)(2,37.88)(3,37.93)(4,37.68)
					};
				\end{axis}
			\end{tikzpicture}
		}
        \end{minipage}
        \label{subfigure2}
    }
    \ref{legendperformance1}
    \caption{Impact of $d_\nu$ and $d_\tau$ on HZMetro.}
    \label{performance1}
\end{figure}
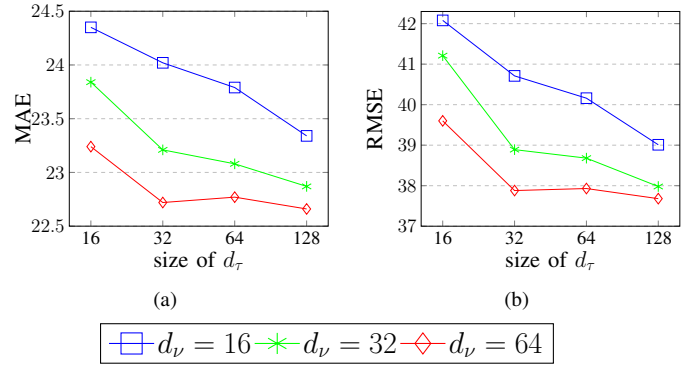

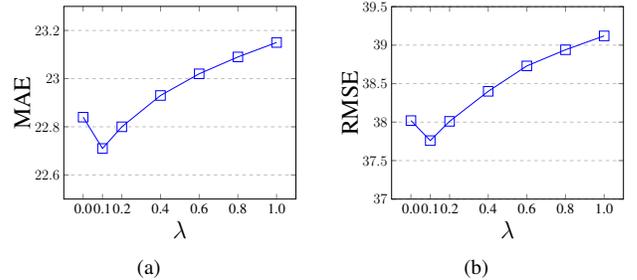
\begin{figure}
\centering
    \subfigure[]{
        \centering
        \begin{minipage}[b]{0.45\columnwidth}
            \resizebox{\linewidth}{!}{
        		\begin{tikzpicture}[scale=0.6]
        			\begin{axis}[
        				xlabel={$\lambda$},
        				ylabel={MAE}, 
        				label style={font=\huge},
        				tick label style = {font=\large},
        				xticklabels={0.0,0.1,0.2,0.4, 0.6, 0.8, 1.0}, 
        				xtick={1,1.5,2,3,4,5,6},
        				ymin=22.50, ymax=23.30,			
        				ymajorgrids=true,
        				grid style=dashed,
        				]
        				\addplot[
        				color=blue,
        				mark=square,
        				mark size=4pt,
        				]
        				coordinates {
        					(1,22.84)(1.5,22.71)(2,22.80)(3,22.93)(4,23.02)(5,23.09)(6,23.15)
        				};
        			\end{axis}
        		\end{tikzpicture}
            }
        \end{minipage}
        \label{lambda:mae}
     }
     \subfigure[]{
        \centering
        \begin{minipage}[b]{0.45\columnwidth}
            \resizebox{\linewidth}{!}{
        		\begin{tikzpicture}[scale=0.6]
        			\begin{axis}[
        				xlabel={$\lambda$},
        				ylabel={RMSE}, 
        				label style={font=\huge},
        				tick label style = {font=\large},
        				xticklabels={0.0,0.1,0.2,0.4, 0.6, 0.8, 1.0}, 
        				xtick={1,1.5,2,3,4,5,6},
        				ymin=37, ymax=39.5,			
        				ymajorgrids=true,
        				grid style=dashed,
        				]
        				\addplot[
        				color=blue,
        				mark=square,
        				mark size=4pt,
        				]
        				coordinates {
        					(1,38.02)(1.5,37.76)(2,38.01)(3,38.40)(4,38.73)(5,38.94)(6,39.12)
        				};
        			\end{axis}
        		\end{tikzpicture}
            }
        \end{minipage}
        \label{lambda:rmse}
     }
     \caption{Impact of $\lambda$ on HZMetro.}
     \label{para:lambda}
\vspace{-20pt}
\end{figure}

\subsubsection{Parameter Sensitivity}\label{sec:para}
Fig.~\ref{performance1} and Fig.~\ref{para:lambda} assess the influence our learned time-aware graph and joint loss optimization on the final forecasting performance. Thus, we conduct a parameter study to analyze the impacts of the three key parameters: node embedding dimensionality $d_\nu$, time embedding dimensionality $d_\tau$, and loss weight factor $\lambda$. From Fig.~\ref{performance1}, we can find that the performance continues to improve as the dimensionality increases, except for slight fluctuations at dimensionality $64$ (red line). TGCRN, with a larger node embedding dimensionality and a larger time embedding dimensionality, can contain more information on graph topology and their dynamics but occupies more parameters, which leads to over-fitting and higher computational costs. Thus, a good practice for finding suitable parameters is to consider the trade-off between performance and computation. We return to this in Section~\ref{sec:cost}.

Fig.~\ref{para:lambda} shows an obvious turning point of the polyline around $\lambda =0.1$, which proves the effectiveness of the fact that time discrepancy learning can mutually promote the interpretability of the learned time-aware graph and the performance, but it is not recommended to make a large proportion too large as an auxiliary task.

\subsubsection{Computational Cost}\label{sec:cost}
Table~\ref{tab:cost} reports the scale of parameters and training time per epoch of the graph-based models. The methods with dynamic graph structure modeling (e.g., TGCRN, ESG) incur higher computational costs to capture the dynamic spatial correlations than those using static graph structures. PVCGN imposes a significant computational burden because it combines multiple graphs on graph convolution. Specifically, TGCRN ($d_\nu=64, d_\tau=32$) has four times more parameters than ESG, but it can achieve significant improvements, as presented in Table \ref{Tab:metro}. As discussed in Section \ref{sec:para}, the prediction performance can be further improved when the model capacity increases. Moreover, TGCRN ($d_\nu=16,d_\tau=16$) with a moderate increase achieves MAE $24.35$, RMSE $42.03$, and MAPE $15.31\%$ on average horizons and still outperforms all baselines. The computational overhead and large model size of TGCRN are due to the modeling of spatial correlations at each time step. However, the changes in correlations between time steps are often small, making it unnecessary to calculate them so frequently. In future work, we will consider how to infer spatial correlations only when crucial changes occur.

\begin{table}[t]
\centering
\caption{Cost computation.}
\label{tab:cost}
\begin{tabular}{@{}ccc@{}}
\toprule
Model                                                & \# Parameters & \begin{tabular}[c]{@{}c@{}}Training Time\\ (per epoch)\end{tabular} \\ \midrule
DCRNN                                                & 373,378        & 2.1s                                                            \\ \midrule
AGCRN                                                & 750,120        & 1.43s                                                           \\ \midrule
GraphWaveNet                                         & 367,396        & 1.3965s                                                          \\ \midrule
PVCGN                                                & 37,598,785      & 48.79s                                                          \\ \midrule
ESG                                                  & 3,936,334       & 7.2461s                                                         \\  \midrule
\begin{tabular}[c]{@{}c@{}}TGCRN\\ $d_\nu=16,d_\tau =16$\end{tabular} & 5,557,331       & 8.62s                                          \\ \midrule
\begin{tabular}[c]{@{}c@{}}TGCRN\\ $d_\nu=64,d_\tau =32$\end{tabular} & 16,675,299      & 10.14s                                          \\
\bottomrule
\end{tabular}
\vspace{-20pt}
\end{table}

\begin{figure*}[t!]
    \centering
    \subfigure[Visualizations of learned time-aware adjacency matrices (Top) and OD passenger transfer flows (Bottom) in time span 08:00 -- 8:15 among stations 3, 4, 5, and 69.]{
        \begin{minipage}[b]{2.0\columnwidth}
            \resizebox{\linewidth}{!}{
                \includegraphics[]{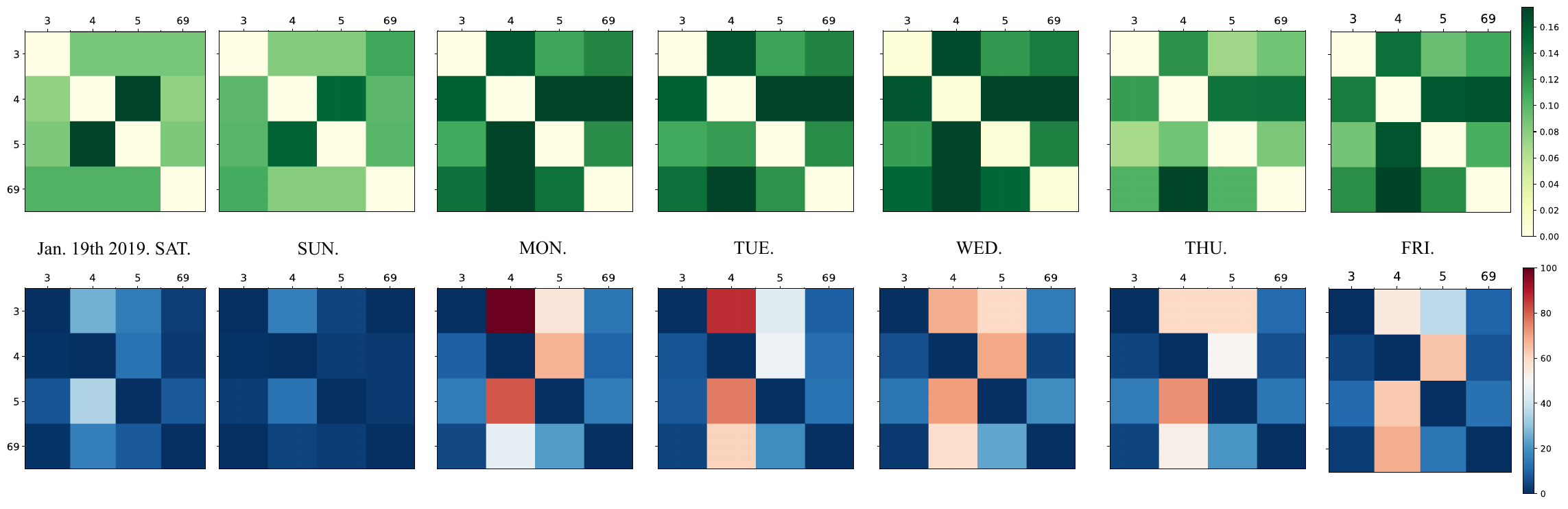}
            }
            \label{case: case1}
        \end{minipage}
    }
    \subfigure[Visualizations of learned time-aware adjacency matrices (Left) and OD passenger transfer flows (Right) over consecutive time spans 08:00 to 09:00]{
        \begin{minipage}[b]{2.0\columnwidth}
            \resizebox{\linewidth}{!}{
                \includegraphics[]{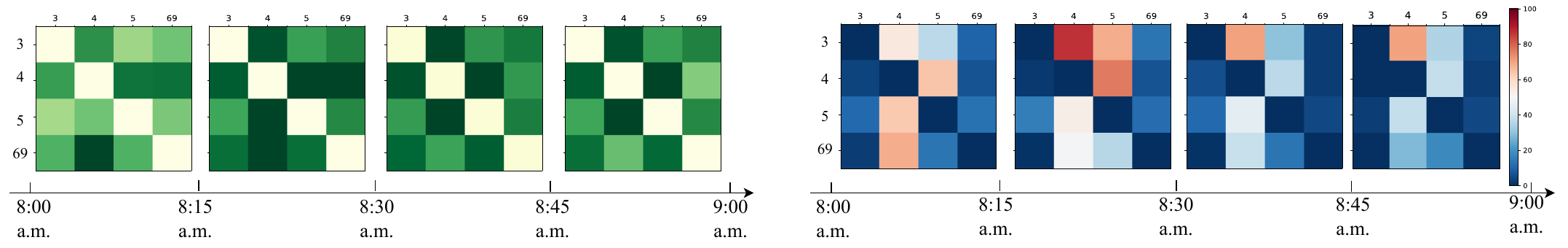}
            }
            \label{case: case2}
        \end{minipage}
    }
    \caption{Visualization of a series of learned adjacency matrices on HZMetro, showing periodicities and trends of time-aware spatial correlations.}
    \label{fig:case_study}
\end{figure*}

\subsection{Visualization}\label{EXP:TagSL}
\subsubsection{Spatial Correlation}\label{EXP:TagSL1}
To observe visually whether the learned time-aware graphs conform to the desired periodicities and trends of spatial correlations (\textbf{Q3}), as discussed in Section~\ref{section:TagSL}, we visualize the learned spatial correlations and time representation.

First, we select four stations and their data in the 08:00 -- 08:15 range from January 19th to 25th, 2019 from the testing dataset of HZMetro. Then we obtain the adjacency matrices at the corresponding timestamps and visualize them as heat maps by enlarging the matrix weights tenfold for highlighting the continuous variations. For the learned spatial correlations, the darker the color, the stronger the correlation. For OD transfer-based correlation, the warmer color, the stronger the association, and the cooler color, the weaker the association. Fig.~\ref{case: case1} shows the learned graphs follow distinct weekday and weekend patterns and are consistent with the OD transfer-based correlations, where there is more demand for metro travel on weekday mornings. 

Moreover, we derive the learned adjacency matrices and OD transfer flows from 08:00 to 09:00 on 24th January 2019 (Thursday). Fig.~\ref{case: case2} shows slight dynamics over consecutive time spans of learned spatial correlations, which has a similar trend as the passenger transfer.

\subsubsection{Time Representation}\label{EXP:TagSL2}
\begin{figure}
	\centering
	\subfigure[w/o Time Discrepancy Learning]{
		\begin{minipage}[b]{0.45\columnwidth}
			\includegraphics[width=1\textwidth]{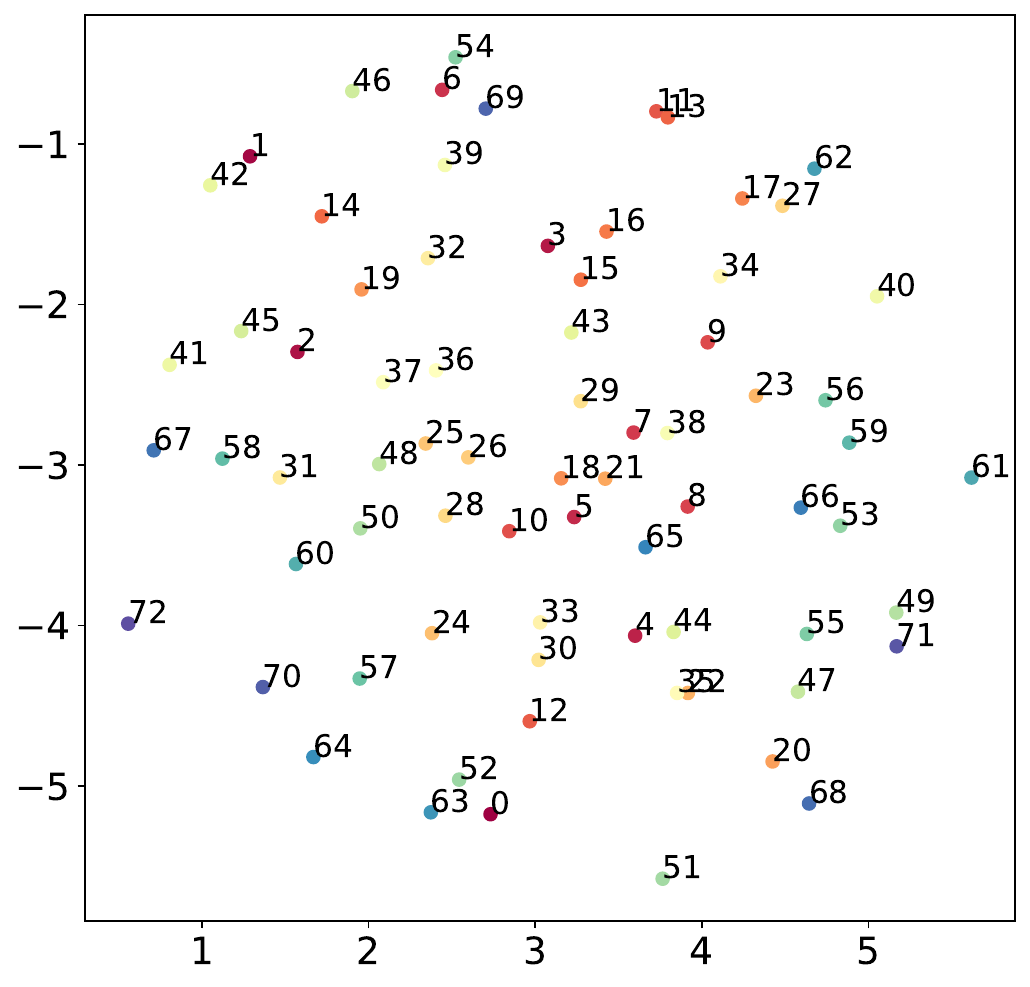}
		\end{minipage}
		\label{fig:vis_no_tdl}
	}
        \subfigure[with Time Discrepancy Learning]{
            \begin{minipage}[b]{0.45\columnwidth}
                \includegraphics[width=1\textwidth]{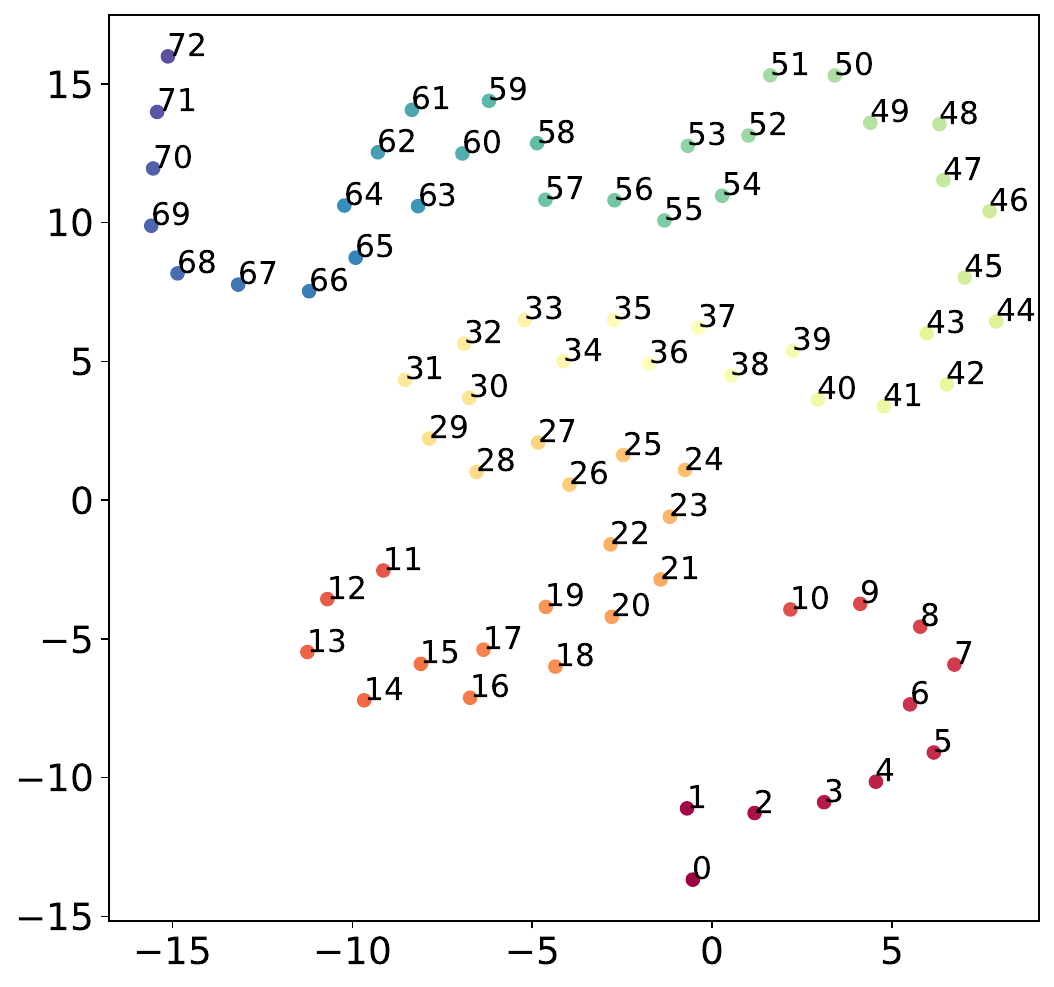}
            \end{minipage}
            \label{fig:vis_with_tdl}
        }
	\caption{Visualization of 2-dimensional time representations w and w/o Time Discrepancy Learning using t-SNE algorithm. The number of timestamps in 15-minute intervals from 05:30 to 11:30 is 73.}
\label{fig:compar_tdl}
\vspace{-20pt}
\end{figure}

To observe the effect of the proposed Time Discrepancy Learning (\textbf{Q4}), we visualize the time representations with and without Time Discrepancy Learning. To do so, we reduce the dimensionality of the time embedding weights of the trained TGCRN from 64 to 2 using t-SNE~\cite{sne}. Fig.~\ref{fig:vis_with_tdl} shows that the representations of time nodes from 0 to 72 exhibit a positional ordering in 2D space with a clear proportional discrepancy, which indicates the effectiveness of the Time Discrepancy Learning module. In contrast, the representations of time nodes without any constraints yield a confusing pattern, as shown in Fig.~\ref{fig:vis_no_tdl}.
 
\section{Related Work}
Spatio-temporal forecasting extends time series forecasting to encompass also a spatial aspect and has attracted substantial attention. Early studies treat this problem as a task of forecasting multiple univariate time series. Then followed statistical methods, e.g., Autoregressive Integrated Moving Average (ARIMA)~\cite{box2015time}, Vector Autoregressive (VAR), and Hidden Markov Models (HMMs). These are linear methods that smooth historical information to predict future state, but they disregard correlations between time series. Researchers have built handcrafted features and then utilize traditional machine learning models, e.g., Linear Regression~\cite{tong2017simpler} or Support Vector Regression~\cite{evgeniou2000regularization},  to capture spatio-temporal dependencies between time series. However, such approaches rely heavily on complex feature engineering to obtain good forecasting performance. This approach is thus constrained by the available domain knowledge and the linear feature representation. More recently, deep learning with powerful non-linear capabilities has become used widely in spatio-temporal forecasting. ConvLSTM~\cite{ConvLSTM} combines LSTMs and CNNs to extract long-term temporal dependencies and spatial relationships among local regions and achieves great performance at the precipitation nowcasting problem. To capture multi-scale temporal dependencies, ST-ResNet~\cite{zhang2017deep} utilizes CNN-based networks to jointly extract spatial and temporal correlations. Other hybrid network-based methods~\cite{lai2018modeling, wang2020multi} also employ CNNs to capture spatial interactions by representing them as global hidden state, but these methods struggle to explicitly model spatial correlations among series.

Recently, GCNs have been leveraged to enable the modeling of hidden dependencies among nodes in graph-structured data. GCNs perform convolution on graph-structured data and can be categorized into two main directions: Spectral-based GCNs~\cite{SpectralGCN, SpectralGCN2} have a solid mathematical foundation and apply convolution on node state and a normalized Laplacian matrix in the spectral domain after Graph Fourier Transform, and then reconstruct the node state after filtering by an Inverse Graph Fourier Transform. This direction faces the limitations of domain dependence and has cubic computational complexity. Next, spatial-based GCNs~\cite{GCN, MPNN} recursively aggregate the representations of the neighbors of a node to update the node's representation in a message-passing manner. Motivated by the flexibility and efficiency of spatial-based GCNs, many spatio-temporal forecasting studies utilize these to capture spatial dependencies between time series, where the graph structure plays an important role in providing topological information. 

There are two prevailing graph structure types: pre-defined graphs and self-learning graphs, according to how the graphs are constructed. Generally, pre-defined graph structures are constructed from domain knowledge and maintain fixed weights during model training and testing~\cite{li2017diffusion, bai2019stg2seq, liu2020physical}. To capture implicit spatial correlations and to contend with scenarios without pre-defined graphs, the self-learning-based methods~\cite{bai2020adaptive, wu2019graph} learn optimized adjacency matrices derived from node embeddings for downstream predictive tasks using a metric function such as inner product similarity. Both pre-defined and self-learning graphs are static during testing and cannot capture dynamic spatial correlations. We also note that~\citet{ye2022learning} employ a neural network-based module that takes the hidden state of time series and builds a series of evolving graph structures. However, this proposal lacks explicit consideration of the periodicities and trends of spatial correlations. In contrast, \textit{TGCRN} not only considers the representations of nodes and time; it also identifies periodicities based on the hidden state of series, and it thus can model dynamic correlations over time that exhibit spatial trends and periodicities.

\section{Conclusion}
We present TGCRN, a novel framework for the forecasting of spatially correlated time series. To the best of our knowledge, this is the first study that takes into account dynamics with periodicities and trends of spatially correlated time series for the purpose of time series forecasting. We proposed an effective method, called time-aware graph structure learning, to exploit time-related regular inter-variable correlations that are represented as a graph structure. We propose GCGRU to jointly capture dynamic spatial and temporal dependencies. Finally, we developed a unified framework with an encoder-decoder architecture that integrates the proposed graph structure learning and GCGRU to output multi-step forecasts. Experiments conducted on several real-world datasets demonstrated that TGCRN is capable of outperforming thirteen existing proposals in terms of forecasting performance.

\section{Acknowledge}
This research was supported by the National Natural Science Foundation of China (Nos. 62176221, 62276215).

\balance
\bibliographystyle{plainnat}
\bibliography{sample}
\end{document}